\documentclass[sigconf]{acmart}

\usepackage{booktabs} 

\usepackage{url}
\usepackage{graphicx}
\usepackage{amsmath}
\usepackage{amsfonts}
\usepackage{amssymb}
\usepackage{dsfont}
\usepackage{mathtools}
\usepackage{makecell}
\usepackage{xcolor,colortbl}
\usepackage{subfigure}
\usepackage{tabularx}
\usepackage{todonotes}
\usepackage[shortlabels,inline]{enumitem}

\newcommand{\vtheta}{\boldsymbol{\theta}}
\newcommand{\vTheta}{\boldsymbol{\Theta}}
\newcommand{\citeinline}[1]{\citeauthor{#1}~\cite{#1}}

\usepackage{array}
\newcolumntype{L}[1]{>{\raggedright\let\newline\\\arraybackslash\hspace{0pt}}m{#1}}

\begin{document}

\copyrightyear{2018}
\acmYear{2018}
\setcopyright{acmlicensed}
\acmConference[KDD '18]{The 24th ACM SIGKDD International Conference on Knowledge Discovery \& Data Mining}{August 19--23, 2018}{London, United Kingdom}
\acmBooktitle{KDD '18: The 24th ACM SIGKDD International Conference on Knowledge Discovery \& Data Mining, August 19--23, 2018, London, United Kingdom}
\acmPrice{15.00}
\acmDOI{10.1145/3219819.3220058}
\acmISBN{978-1-4503-5552-0/18/08}

\title{Hyperparameter Importance Across Datasets}

\author{Jan N. van Rijn}
\affiliation{%
  \institution{Albert-Ludwigs-Universit\"{a}t Freiburg}
  \streetaddress{Georges-Köhler-Allee 052}
  \city{Freiburg}
  \state{Germany}
  \postcode{79110}
}
\email{vanrijn@cs.uni-freiburg.de}

\author{Frank Hutter}
\affiliation{%
  \institution{Albert-Ludwigs-Universit\"{a}t Freiburg}
  \streetaddress{Georges-Köhler-Allee 052}
  \city{Freiburg}
  \state{Germany}
  \postcode{79110}
}
\email{fh@cs.uni-freiburg.de}

\begin{abstract}
With the advent of automated machine learning, automated hyperparameter optimization methods are by now routinely used in data mining. However, this progress is not yet matched by equal progress on automatic analyses that yield information beyond performance-optimizing hyperparameter settings.
In this work, we aim to answer the following two questions:  
Given an algorithm, what are generally its most important hyperparameters, and what are typically good values for these?
We present methodology and a framework to answer these questions based on meta-learning across many datasets. 
We apply this methodology using the experimental meta-data available on OpenML to determine the most important hyperparameters of support vector machines, random forests and Adaboost, and to infer priors for all their hyperparameters. 
The results, obtained fully automatically, provide a quantitative basis to focus efforts in both manual algorithm design and in automated hyperparameter optimization.
The conducted experiments confirm that the hyperparameters selected by the proposed method are indeed the most important ones and that the obtained priors also lead to statistically significant improvements in hyperparameter optimization. 
\end{abstract}

%
%



\maketitle

\section{Introduction}

The performance of modern machine learning and data mining methods highly depends on their hyperparameter settings. As a consequence, there has been a lot of recent work and progress on hyperparameter optimization, with methods including random search~\cite{Bergstra2012}, Bayesian optimization~\cite{Bergstra2011,Hutter2011,Snoek2012,Swersky2013,Klein2017}, evolutionary optimization~\cite{Loshchilov2016}, meta-learning~\cite{Brazdil2008,Gomes2012,Reif2012,Miranda2014,Rijn2015,Rijn2016} and bandit-based methods~\cite{Jamieson2016,Li2017}.

Based on these methods, it is now possible to build reliable automatic machine learning (AutoML) systems~\cite{Thornton2013,Feurer2015}, which -- given a new dataset $\mathcal{D}$ -- determine a custom combination of algorithm and hyperparameters that performs well on $\mathcal{D}$. 
However, this recent rapid progress in hyperparameter optimization and AutoML carries a risk with it: if researchers and practitioners rely exclusively on automated methods for finding performance-optimizing configurations, they do not obtain any intuition or information beyond the single configuration chosen.
To still provide such intuition in the age of automation, we advocate the development of automated methods that provide high-level insights into an algorithm's hyperparameters, based on a wide range of datasets. 

When using a new algorithm on a given dataset, it is typically a priori unknown which hyperparameters should be tuned, what are good ranges for these, and which values in these ranges are most likely to yield high performance. 
Currently these decisions are typically made based on a combination of intuition about the algorithm and trial \& error.
While various post-hoc analysis techniques exist that, for a given dataset and algorithm, determine what were the most important hyperparameters and which of their values tended to yield good performance, in this work we study the same question across many datasets.
For many well-known algorithms, there already exists some intuition about which hyperparameters impact performance most. For example, for support vector machines, it is commonly believed that the gamma and complexity hyperparameters are most important, and that a certain trade-off exists between these two. However, the empirical evidence for this is limited to a few datasets and therefore rather anecdotal. 

In this work, given an algorithm, we aim to answer the following two questions:  
\begin{enumerate}
\item Which of the algorithm's hyperparameters matter most for empirical performance?
\item Which values of these hyperparameters are likely to yield high performance?
\end{enumerate}
We will introduce methods to answer these questions across datasets and demonstrate these methods for three commonly used classifiers: support vector machines (SVMs), random forests and Adaboost. 
Specifically, we apply the post-hoc analysis technique of functional ANOVA~\cite{Hutter2014} to each of the aforementioned classifiers on a wide range of datasets, drawing on the experimental data available on OpenML~\cite{Vanschoren2014}. 
Using the same available experimental data, we also infer prior distributions over which hyperparameter values work well. 
Several experiments demonstrate that the trends we find (about which hyperparameters tend to be important and which values tend to perform well) generalize to new datasets.

Our contributions are as follows:
\begin{enumerate}
\item We present a methodology and a framework that leverage functional ANOVA to study hyperparameter importance across datasets. 
\item We apply this to analyze the importance of SVMs, random forests and Adaboost on $100$ datasets from OpenML, and confirm that the hyperparameters determined as the most important ones indeed are the most important ones to optimize.
\item Using the same experimental data, we infer priors over which values of these hyperparameters perform well and confirm that these priors yield statistically significant improvements for a modern hyperparameter optimization method.
\item In order to make this study reproducible, all experimental data is made available on OpenML. The results of all analyses are available in a separate Jupyter Notebook. 
\item Overall, this work is the first to provide quantitative evidence for which hyperparameters are important and which values should be considered, providing a better scientific basis for the field than previous knowledge based mainly on intuition. 
\end{enumerate}

The remainder of this paper is organized as follows. 
In Section~\ref{sec:related} we position our contributions with respect to similar works in the field. 
Section~\ref{sec:background} covers relevant background information about functional ANOVA.
Section~\ref{sec:methods} formally introduces the methods that we propose, and Section~\ref{sec:algorithms} defines the algorithms and hyperparameters upon which we apply them. 
We then conduct two experiments: 
Section~\ref{sec:importance} covers the experiments that show which hyperparameters are important across datasets; and Section~\ref{sec:priors} covers the experiments that show how to use the experimental data on OpenML to infer good priors. 
Section~\ref{sec:conclusions} concludes.

\section{Related Work}
\label{sec:related}
We review related work on hyperparameter importance and priors. 

\paragraph{\bf Hyperparameter Importance.} 
Various techniques exists that allow for the assessment of hyperparameter importance. 
\citeinline{Breiman2001} showed in his seminal paper how random forests can be used to assess attribute importance: if removing an attribute from the dataset yields a drop in performance, this is an indication that the attribute was important. 
\emph{Forward selection}~\cite{Hutter2013} is based on this principle. It predicts the performance of a classifier based on a subset of hyperparameters that is initialized empty and greedily filled with the next most important hyperparameter. 
\emph{Ablation Analysis}~\cite{Fawcett2016,Biedenkapp2017} requires a default setting and an optimized setting and calculates a so-called ablation trace, which embodies how much the hyperparameters contributed towards the difference in performance between the two settings. 
\emph{Functional ANOVA} (as explained in detail in the next section) is a powerful framework that can detect the importance of both individual hyperparameters and interaction effects between arbitrary subsets of hyperparameters. 
Although all of these methods are very useful in their own right, none of these has yet been applied to analyze hyperparameters across datasets. We will base our work in this realm on functional ANOVA since it is computationally far more efficient than forward selection, can detect interaction effects, and (unlike ablation analysis) does not rely on a specific default configuration. The proposed methods are, however, by no means limited to functional ANOVA.

In a preliminary study, we already reported on important hyperparameters of random forests and Adaboost~\cite{Rijn2017}. 

\paragraph{\bf Priors.} 
The field of meta-learning (e.g.,~\citeinline{Brazdil2008}) is implicitly based on priors: a model is trained on data characteristics (so-called meta-features) and performance data from similar datasets, and the resulting predictions are used to recommend a configuration for the dataset at hand. These techniques have been successfully used to recommend good hyperparameter settings~\cite{Soares2004,Miranda2014}, to warm-start optimization procedures~\cite{Feurer2015a} or prune search spaces~\cite{Wistuba2015}. However, it is hard to select an adequate set of meta-features. Moreover, obtaining good meta-features comes at the cost of run time. 
This work can be seen as an alternative approach to meta-learning that does not require the aforementioned meta-features. 

Multi-task Bayesian optimization~\cite{Swersky2013} offers a different approach to meta-learning that alleviates meta-features. 
A multi-task model (typically a Gaussian Process~\cite{Bonilla2008}) is fitted on the outcome of classifiers to determine correlations between tasks, which can be exploited for hyperparameter optimization on a new task. 
However, this approach suffers from the cubic complexity of Gaussian processes. While a recent more scalable alternative for multi-task Bayesian optimization is to use Bayesian neural networks~\cite{Springenberg2016}, to the best of our knowledge, this approach has not been evaluated at large scale yet.

The class of Estimation of Distribution (EDA) algorithms (e.g.~\citeinline{Larraanaga2001}) optimizes a given function by iteratively fitting a probability distribution to points in the input space with high performance and using this probability distribution as a prior to sample new points from. 
Drawing on this, the method we propose determines priors over good hyperparameter values by using hyperparameter performance data on different datasets.

\section{Background: Functional ANOVA}
\label{sec:background}

The functional ANOVA framework for analyzing the importance of hyperparameters introduced by \citeinline{Hutter2014} is based on a regression model that yields predictions $\hat{y}$ for the performance of arbitrary hyperparameter settings. It determines how much each hyperparameter (and each combination of hyperparameters) contributes to the variance of $\hat{y}$ across the algorithm's hyperparameter space $\vTheta$. Since we will use this technique as part of the proposed method, we now discuss it in more detail.

\paragraph{\bf Notation.} 
Algorithm $A$ has $n$ hyperparameters with domains $\Theta_1, \ldots, \Theta_n$ and \emph{configuration space} $\vTheta = \Theta_1 \times \ldots \times \Theta_n$. Let $N = \{1, \dots, n\}$ be the set of all hyperparameters of $A$.
An instantiation of $A$ is a vector $\vtheta = \langle \theta_1, \ldots, \theta_n \rangle$ with $\theta_i \in \Theta_i$ (this is also called a \emph{configuration} of $A$). A partial instantiation of $A$ is a vector $\vtheta_U = \langle \theta_i, \ldots, \theta_j \rangle$ with a subset $U \subseteq N$ of the hyperparameters fixed, and the values for other hyperparameters unspecified. (Note that from this it follows that $\vtheta_N = \vtheta$). 

 \begin{figure*}[tb]
   \begin{center}
     \subfigure[Gamma] {
       \includegraphics[width=.32\textwidth]{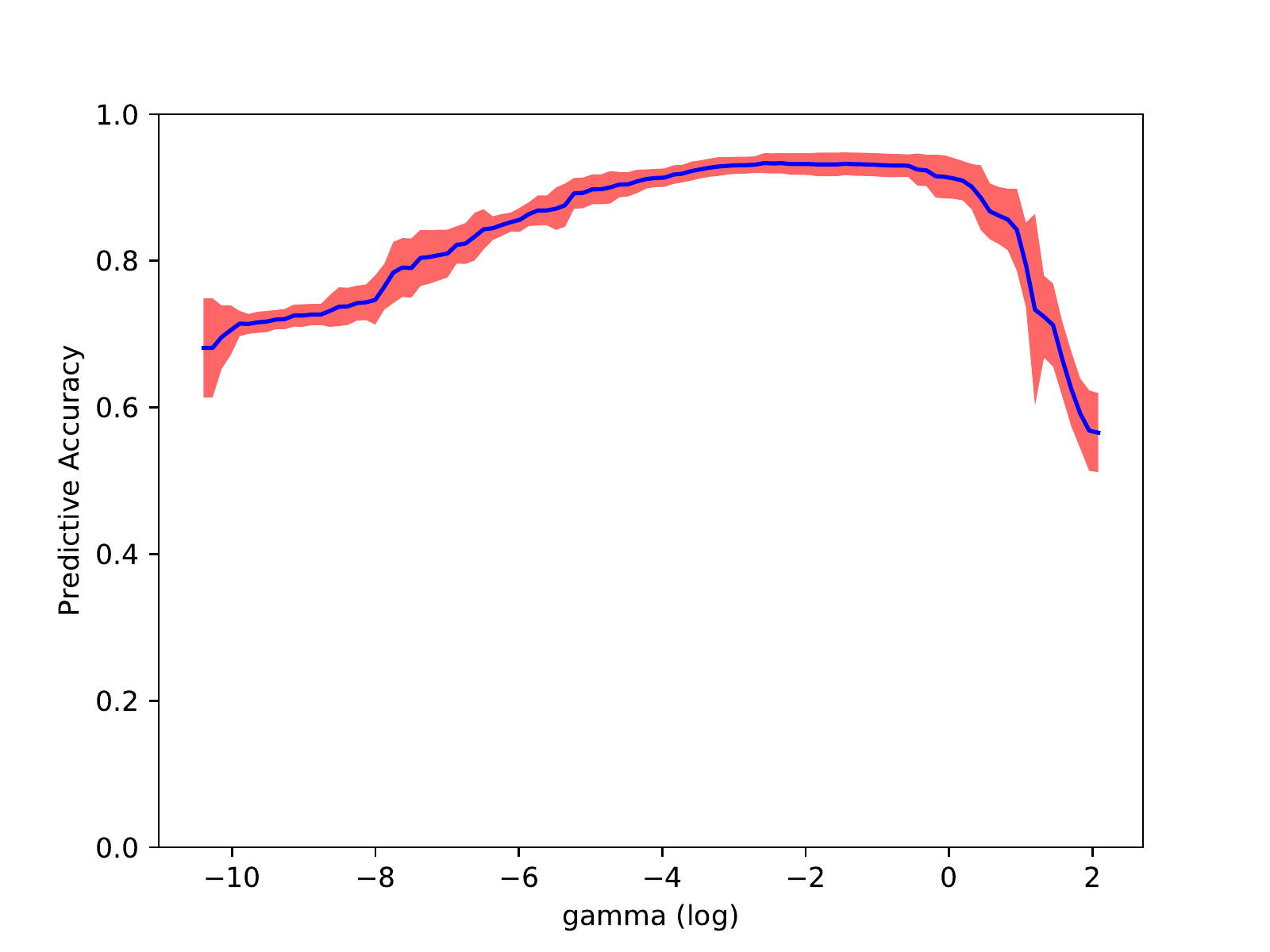}
     }
     \subfigure[Complexity] {
       \includegraphics[width=.32\textwidth]{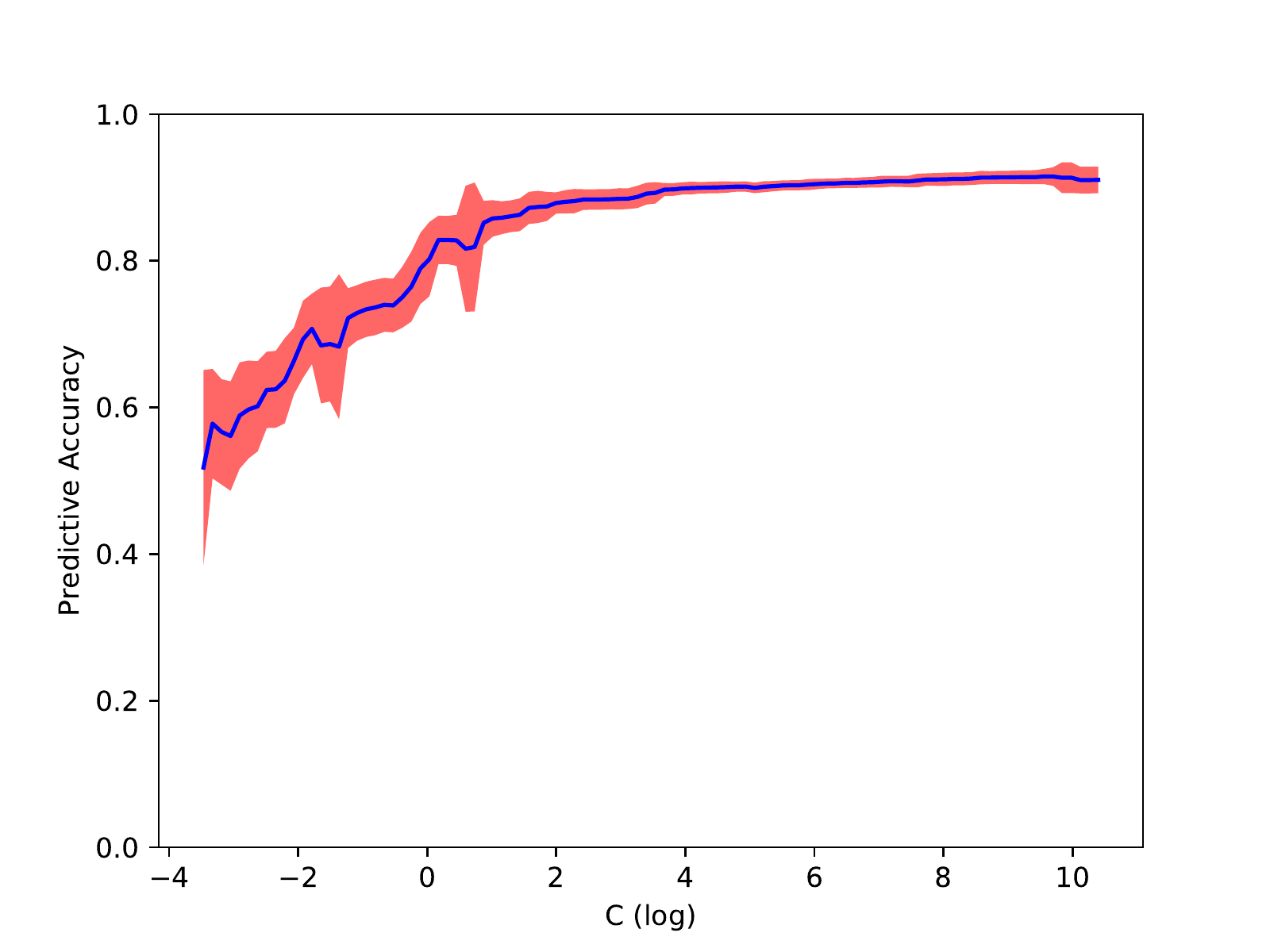}
     }
     \subfigure[Gamma vs. Complexity] {
       \includegraphics[width=.32\textwidth]{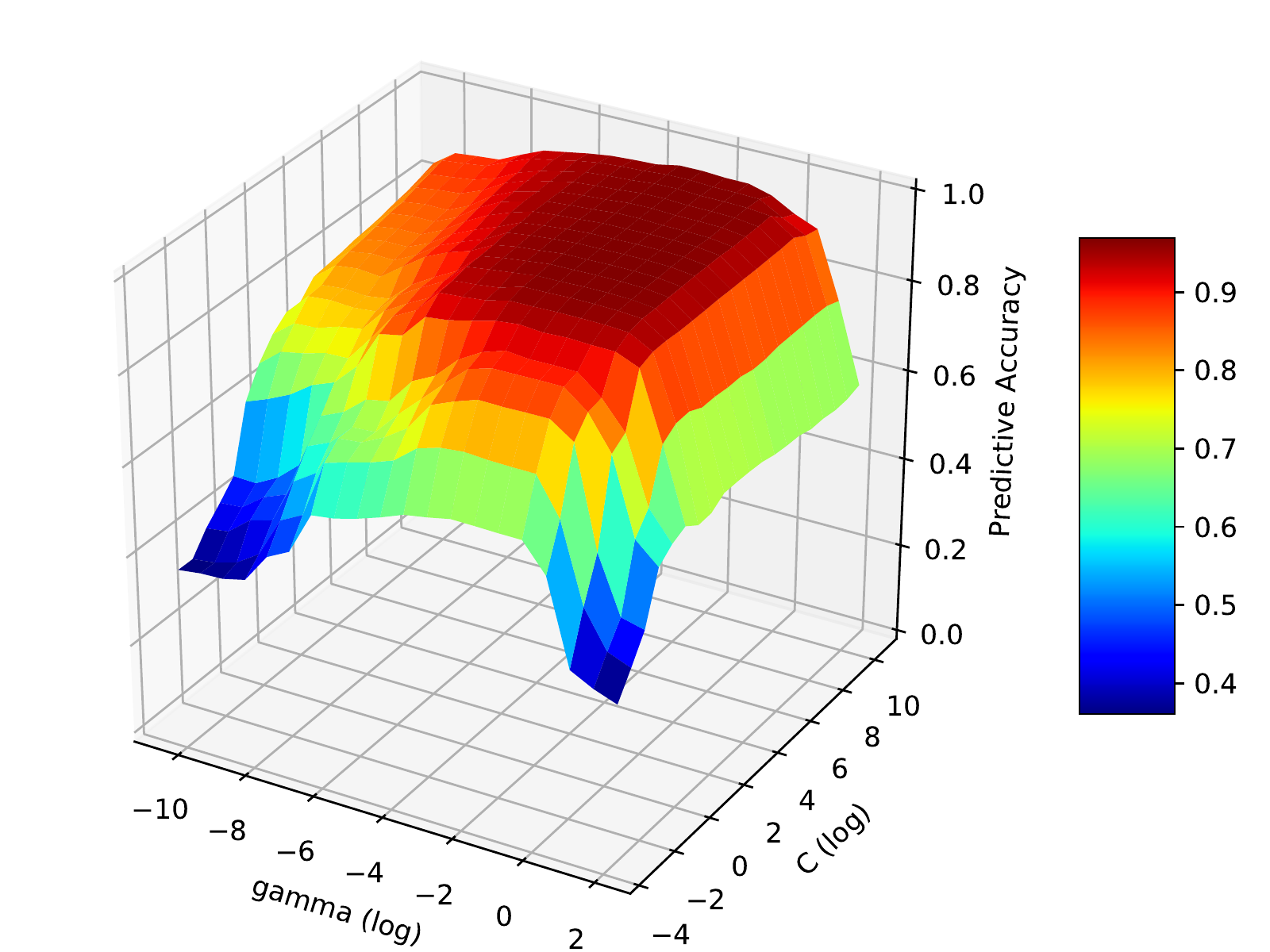}
     }
     \caption{Marginal predictions for a SVM with RBF kernel on the letter dataset. The hyperparameter values are on a log scale. \label{fig:marginal}}
   \end{center}
 \end{figure*}

\paragraph{\bf{Efficient marginal predictions.}} 
The \emph{marginal performance} $\hat{a}_U(\vtheta_U)$ is defined as the average performance of all complete instantiations $\vtheta$ that agree with $\vtheta_U$ in the instantiations of hyperparameters $U$. 
To illustrate the concept of marginal predictions, Figure~\ref{fig:marginal} shows marginal predictions for two hyperparameters of SVMs and their union. 
We note that such marginals average over \emph{all} instantiations of the hyperparameters not in $U$, and as such depend on a very large number of terms (even for finite hyperparameter ranges, this number of terms is exponential in the remaining number of hyperparameters $N \setminus U$). 
However, for the predictions $\hat{y}$ of a tree-based model, the average over these terms can be computed exactly by a procedure that is linear in the number of leaves in the model~\cite{Hutter2014}. 

\paragraph{\bf{Functional ANOVA.}} Functional ANOVA~\cite{Sobol1993,Huang1998,Jones1998,Hooker2007} decomposes a function $\hat{y}:\Theta_1\times\cdots\times\Theta_n \rightarrow \mathds{R}$ into additive components that only depend on subsets of the hyperparameters $N$:
\begin{equation}
  \hat{y}(\vtheta) = \sum_{U\subseteq N} \hat{f}_{U}(\vtheta_U)
\end{equation}

\noindent{}The components $\hat{f}_{U}(\vtheta_U)$ are defined as follows:
\begin{equation}
\label{eq:f_u}\hat{f}_{U}(\vtheta_U)=\begin{cases}
		\hat{f}_\emptyset & \!\!\!\!\text{if $U=\emptyset$}.\!\!\\
    \hat{a}_{U}(\vtheta_U) - \sum_{W \subsetneq U} \hat{f}_{W}(\vtheta_W) & \!\!\!\!\text{otherwise},\!\!\
  \end{cases}
\end{equation}
\noindent{}where the constant $\hat{f}_\emptyset$ is the mean value of the function over its domain.
Our main interest is the result of the unary functions $\hat{f}_{\{j\}}(\vtheta_{\{j\}})$, which capture the effect of varying hyperparameter $j$, averaging across all possible values of all other hyperparameters. 
Additionally, the functions $\hat{f}_U(\vtheta_U)$ for $|U|>1$ capture the interaction effects between all variables in $U$ (excluding effects of subsets $W \subsetneq U$).

Given the individual components, functional ANOVA decomposes the variance $\mathds{V}$ of $\hat{y}$ into the contributions by all subsets of hyperparameters $\mathds{V}_U$:
\begin{eqnarray}
\mathds{V} = \sum_{U\subset N} \mathds{V}_U, \;\; \text{with} \;\; \mathds{V}_U = \frac{1}{||\vTheta_U||} \int \hat{f}_U(\vtheta_U)^2  d\vtheta_U, \label{eqn:fanova_importance}
\end{eqnarray}
where $\frac{1}{||\vTheta_U||}$ is the probability density of the uniform distribution across $\vTheta_U$.

To apply functional ANOVA, we first collect performance data $\langle{}\vtheta_i,y_i\rangle{}_{k=1}^{K}$ that captures the performance $y_i$ (e.g., accuracy or AUC score) 
of an algorithm $A$ with hyperparameter settings $\vtheta_i$. We then fit a random forest model to this data and use functional ANOVA to decompose the variance of each of the forest's trees $\hat{y}$ into contributions due to each subset of hyperparameters. 
Importantly, based on the fast prediction of marginal performance available for tree-based models, this is an efficient operation requiring only seconds in the experiments for this paper. 
Overall, based on the performance data $\langle{}\vtheta_i,y_i\rangle{}_{k=1}^{K}$, functional ANOVA thus provides us with the relative variance contributions of each individual hyperparameter (with the relative variance contributions of all subsets of hyperparameters summing to one).

This leads to the notion of hyperparameter importance. 
When a hyperparameter is responsible for a large fraction of the variance, setting this hyperparameter correctly is important for obtaining good performance, and it should be tuned properly. 
When a hyperparameter is not responsible for a lot of variance, it is deemed less important. 

Besides attributing the variance to single hyperparameters, functional ANOVA also determines the interaction effects of sets of hyperparameters.
This potentially gives insights in which hyperparameters can be tuned independently and which are dependent on each other and should thus be tuned together.
In the hypothetical case where there are no interaction effects between any of the hyperparameters, all hyperparameters could be tuned individually by means of a simple hill-climbing algorithm.

By design, functional ANOVA operates on the result of a single hyperparameter optimization procedure on a single dataset. 
This leaves room for questions, such as: 
\begin{enumerate*}[(i)]
  \item Which hyperparameters are important in general?
  \item Are the same hyperparameters often important, or does this vary per dataset?
  \item Given a new dataset, on which a hyperparameter procedure is to be ran, which hyperparameters should be optimized and what are sensible ranges?
\end{enumerate*}
We will investigate these questions in the next section. 

\section{Methods}
\label{sec:methods}

We address the following problem. Given 
\begin{itemize}
    \item an algorithm with configuration space $\vTheta$
    \item a large number of datasets $\mathcal{D}^{(1)}, \dots, \mathcal{D}^{(M)}$, with $M$ being the number of datasets
    \item for each of the datasets, a set of empirical performance measurements $\langle{}\vtheta_i,y_i\rangle{}_{i=1}^{K}$ for different hyperparameter settings $\vtheta_i \in \vTheta$, 
\end{itemize}
we aim to determine which hyperparameters affect the algorithm's empirical performance most, and which values are likely to yield good performance.

\subsection{Important Hyperparameters}

Current knowledge about hyperparameter importance is mainly based on a combination of intuition, own experience and folklore knowledge. 
To instead provide a data-driven quantitative basis for this knowledge, in this section we introduce methodology for determining which hyperparameters are generally important, measured across datasets. 

\paragraph{\bf Determining Important Hyperparameters.}
For a given algorithm $A$ and a given dataset, we use the performance data $\langle{}\vtheta_i,y_i\rangle{}_{i=1}^{K}$ collected for $A$ on this dataset to fit functional ANOVA's random forests to. Functional ANOVA then returns the variance contribution $\mathds{V}_{j} / \mathds{V}$ of every hyperparameter $j\in N$, with high values indicating high importance. 
We then study the distribution of these variance contributions across datasets to obtain empirical data regarding which hyperparameters tend to be most important. 

It is possible that a given set of hyperparameters is responsible for a high variance on many datasets, but the best performance is typically achieved with the same set of values.
We note that this method will flag such hyperparameters as important,
although it could be argued that they have appropriate defaults and do not need to be tuned.
Whether this is the case can be determined by various procedures, for example the one introduced in Section~\ref{sec:methods_priors}. 

For some datasets, the measured performance values $y_i$ are constant, indicating that none of the hyperparameters are important; we therefore removed these datasets from the respective experiments.

\paragraph{\bf Verification.}
Functional ANOVA uses a mathematically clearly defined quantity ($\mathds{V}_{j} / \mathds{V}$) to define a hyperparameter's importance, but it is important to verify whether this agrees with other, potentially more intuitive, notions of hyperparameter importance. 
To confirm the results of functional ANOVA, we therefore propose to verify in an expensive, post-hoc analysis to what extent its results align with an intuitive notion of how important a hyperparameter is in hyperparameter optimization.

One intuitive way to measure the importance of a hyperparameter $\theta$ is to assess the performance obtained in an optimization process that leaves $\theta$ fixed.
However, similar to ablation analysis~\cite{Fawcett2016}, the outcome of this approach depends strongly on the value that $\theta$ is fixed to; e.g., fixing a very important hyperparameter to a good default value would result in labelling it not important (this indeed happened in various cases in our preliminary experiments). To avoid this problem and to instead quantify the importance of setting $\theta$ to a good value in its range, we perform $k$ runs of the optimization process of all hyperparameters but $\theta$, each time fixing $\theta$ to a different value spread uniformly over its range; in the end, we average the results of these $k$ runs. Leaving out an important hyperparameter $\theta$ is then expected to yield worse results than leaving out an unimportant hyperparameter $\theta'$.
As a hyperparameter optimization procedure for this verification procedure, we simply use random search, to avoid any biases.

Formally, for each hyperparameter $\theta_j$ we measure $y^*_{j,f}$ as the result of a random search for maximizing accuracy, fixing $\theta_j$ to a given value $f \in F_j$. (For categorical $\theta_j$ with domain $\Theta_j$, we used $F_j=\Theta_j$; for numeric $\theta_j$, we set $F_j$ to a set of $k = 10$ values spread uniformly over $\theta_j$'s range.) %
We then compute $y^*_j = \frac{1}{|F_j|}\sum_{f \in F_j}y^*_{j,f}$, representing the score when not optimizing hyperparameter $\theta_j$, averaged over fixing $\theta_j$ to various values it can take. 
Hyperparameters with lower $y^*_j$ are then judged to be more important, since performance deteriorates more when they are set sub-optimally.

\subsection{Priors for Good Hyperparameter Values}
\label{sec:methods_priors}
Knowing what are the important hyperparameters, an obvious next question is what are good values for these hyperparameters. 
These values can be used to define defaults, or to sample from in hyperparameter optimization. 

\paragraph{\bf Determining Useful Priors.}
We aim to build priors based on the performance data observed across datasets. There are several existing methods for achieving this on a single dataset that we drew inspiration from. In hyperparameter optimization, the Tree-structured Parzen Estimator (TPE) by \citeinline{Bergstra2011} keeps track of an algorithm's best observed hyperparameter configurations $\vTheta_{best}$ on a given dataset and for each hyperparameter fits a 1-dimensional Parzen Estimator to the values it took in $\vTheta_{best}$. 
Similarly, as an analysis technique to study which values of a hyperparameter perform well, \citeinline{Loshchilov2016} proposed to fit kernel density estimators (see, e.g.,~\cite{Scott2015}) to these values. 
Here, we follow this latter procedure, but instead of using the hyperparameter configurations that performed well on a given dataset, we used the top $n$ configurations observed for each of the datasets; in our experiments, we set $n=10$.
We only used 1-dimensional density estimators in this work, because the amount of data required to adequately fit these is known to be reasonable. We note that this is merely one possible choice, and that future work could focus on fitting other types of distributions to this data. 

\paragraph{\bf Verification.}
As in the case of hyperparameter importance, we propose an expensive post-hoc analysis to verify whether the priors over good hyperparameter values identified above are useful and generalize across datasets.
Specifically, as a quantifiable notion of usefulness, we propose to evaluate the impact of using the prior distributions defined above within a hyperparameter optimization procedure. 
For this, we use the popular bandit-based hyperparameter optimization method Hyperband~\cite{Li2017}. Hyperband is based on the procedure of successive halving~\cite{Jamieson2016}, which evaluates a large number of randomly-chosen configurations using only a small budget, and iteratively increases this budget,  at each step only retaining a fraction of configurations that are best so far. 
For each dataset, we propose to run two versions of this optimization procedure: one sampling uniformly from the hyperparameter space and one sampling from the obtained priors. 
If the priors are indeed useful and generalize across datasets, the optimizer that uses them should obtain better results on the majority of the datasets.
Of course, for each dataset on which this experiment is performed, the priors should be obtained on empirical performance data that was not obtained from this dataset. 

We note that -- due to differences between datasets -- there are bound to be datasets for which using priors from other datasets deteriorates performance. 
However, since human engineers have successfully used prior knowledge to define typical ranges to consider, our hypothesis is that the data-driven priors will improve the optimization procedure's results on most datasets.

\subsection{Algorithm Performance Data}
The proposed methods do not crucially rely on how exactly the training performance data was obtained. 
We note, however, that for all training datasets the data should be gathered with a wide range of hyperparameter configurations (to allow the construction of predictive performance models for functional ANOVA) and should contain close-to-optimal configurations (to allow the construction of good priors).

We note that for many common algorithms, the open machine learning environment OpenML~\cite{Vanschoren2014} already contains very comprehensive performance data for different hyperparameter configurations on a wide range of datasets. OpenML also defines curated benchmarking suites, such as the OpenML100~\cite{Bischl2017}.
We therefore believe that the proposed methods can in principle be used directly on top of OpenML to automatically provide and refine insights as more data becomes available.

In our experiments, which involve classifiers with up to six hyperparameters, we indeed used data from OpenML. 
We ensured that for each dataset at least $150$ runs with different hyperparameters were available to make functional ANOVA's model reliable enough. 
We generated additional runs for classifiers that did not meet this requirement by executing random configurations on a large compute cluster. 
We note that for larger hyperparameter spaces, more sophisticated data gathering strategies are likely required to accurately model the performance of the best configurations. 

\subsection{Computational Complexity of Analysis Techniques}

While we also propose the use of expensive, post-hoc verification methods to confirm the results of our analysis, we would like to emphasize that the proposed analysis techniques themselves are computationally very efficient. 
Their complexity is dominated by the cost of fitting functional ANOVA's random forest to the performance data observed for each of the datasets. 
The cost of the remainder of functional ANOVA, and of fitting the Gaussian kernel density estimator is negligible. 
In the experiments we conducted, given an algorithm's performance data, performing the analyses required only a few seconds. 

\section{Algorithms and Hyperparameters}
\label{sec:algorithms}
We analyze the hyperparameters of three classifiers implemented in scikit-learn~\cite{Pedregosa2011,Buitinck2013}: random forests~\cite{Breiman2001}, Adaboost (using decision trees as base-classifier)~\cite{Freund1995} and SVMs~\cite{Chang2011}. The SVMs are analysed with two different kernel types: radial basis function (RBF) and sigmoid.

For each of these, to not incur any bias from our choice of hyperparameters and ranges, we used exactly the same hyperparameters and ranges as the automatic machine learning system Auto-sklearn~\cite{Feurer2015}.\footnote{There was one exception: For technical reasons, in random forests, we modelled the maximal number of features for a split as a fraction of the number of available features (with range $[0.1,0.9]$).} 
The hyperparameters, ranges and scales are listed in Tables~\ref{tab:svm}--\ref{tab:adaboost}.

\begin{table*}[t]
  \caption{SVM Hyperparameters.}
  \label{tab:svm}
  \begin{tabular}{L{2.5cm} L{3cm} L{11cm}}
    \hline
    hyperparameter & values & description \\
    \hline
    complexity &  $[2^{-5}, 2^{15}]$ (log-scale) & Soft-margin constant, controlling the trade-off between model simplicity and model fit. \\ 
    coef0 & $[-1, 1]$ & Additional coefficient used by the kernel (sigmoid kernel only). \\
    gamma  & $[2^{-15}, 2^{3}]$ (log-scale) & Length-scale of the kernel function, determining its locality. \\
    imputation & $\{$mean, median, mode$\}$ & Strategy for imputing missing numeric variables. \\
    shrinking & $\{$true, false$\}$ & Determines whether to use the shrinking heuristic (introduced in~\cite{Joachims1999}). \\
    tolerance & $[10^{-5}, 10^{-1}]$ (log-scale) & Determines the tolerance for the stopping criterion. \\
    \hline
  \end{tabular}
\end{table*}

\begin{table*}[t]
  \caption{Random Forest Hyperparameters.\label{tab:randomforest}}
  \begin{tabular}{L{2.5cm} L{3cm} L{11cm}}
    \hline
    hyperparameter & values & description \\
    \hline
    bootstrap & $\{$true, false$\}$ & Whether to train on bootstrap samples or on the full train set. \\ 
    max. features & $[0.1,0.9]$ & Fraction of random features sampled per node.  \\
    min. samples leaf & $[1, 20]$ & The minimal number of data points required in order to create a leaf. \\
    min. samples split & $[2, 20]$ & The minimal number of data points required to split an internal node. \\
    imputation & $\{$mean, median, mode$\}$ & Strategy for imputing missing numeric variables.  \\
    split criterion & $\{$entropy, gini$\}$ & Function to determine the quality of a possible split. \\
    \hline
  \end{tabular}
\end{table*}

\begin{table*}[t]
  \caption{Adaboost Hyperparameters.}
  \label{tab:adaboost}
  \begin{tabular}{L{2.5cm} L{3cm} L{11cm}}
    \hline
    hyperparameter & values & description \\
    \hline
    algorithm & $\{$SAMME, SAMME.R$\}$ & Determines which boosting algorithm to use. \\
    imputation & $\{$mean, median, mode$\}$ & Strategy for imputing missing numeric variables. \\
    iterations & $[50, 500]$ & Number of estimators to build. \\
    learning rate &  $[0.01, 2.0]$ (log-scale) & Learning rate shrinks the contribution of each classifier. \\
    max. depth & $[1, 10]$ & The maximal depth of the decision trees. \\
    \hline
  \end{tabular}
\end{table*}

\paragraph{\bf Preprocessing.}
We used the same data preprocessing steps for all algorithms.
Missing values are imputed (categorical features with the mode; for numerical features, the imputation strategy was one of the hyperparameters), categorical hyperparameters are one-hot-encoded, and constant features are removed. As support vector machine's are sensitive to the scale of the input variables, the input variables for the SVM's are scaled to have unit variance.  
Of course, all these operations are performed based on information obtained from the training data.

\paragraph{\bf Datasets.}
We performed all experiments on the datasets from the OpenML100~\cite{Bischl2017}. The OpenML100 is a curated benchmark suite, containing 100 datasets from various domains. The datasets contain between $500$ and $100,\!000$ data points, are generally well-balanced and are all linked to a scientific publication. 
These criteria ensure that the datasets pose a challenging and meaningful classification task, and the results are comparable to earlier studies. 

\section{Hyperparameter Importance}
\label{sec:importance}
We now discuss the results of the experiment for determining the most important hyperparameters per classifier. 
All together, this analysis is based on the performance data of $250,\!195$ algorithm runs over the $100$ datasets using $3,\!184$ CPU days to generate. 
All performance data we used is publicly available on OpenML\footnote{Full details: \url{https://www.openml.org/s/71}}.

\begin{figure}[tb]
  \begin{center}
    \subfigure[Marginal contribution per dataset]{
      \includegraphics[width=.75\columnwidth]{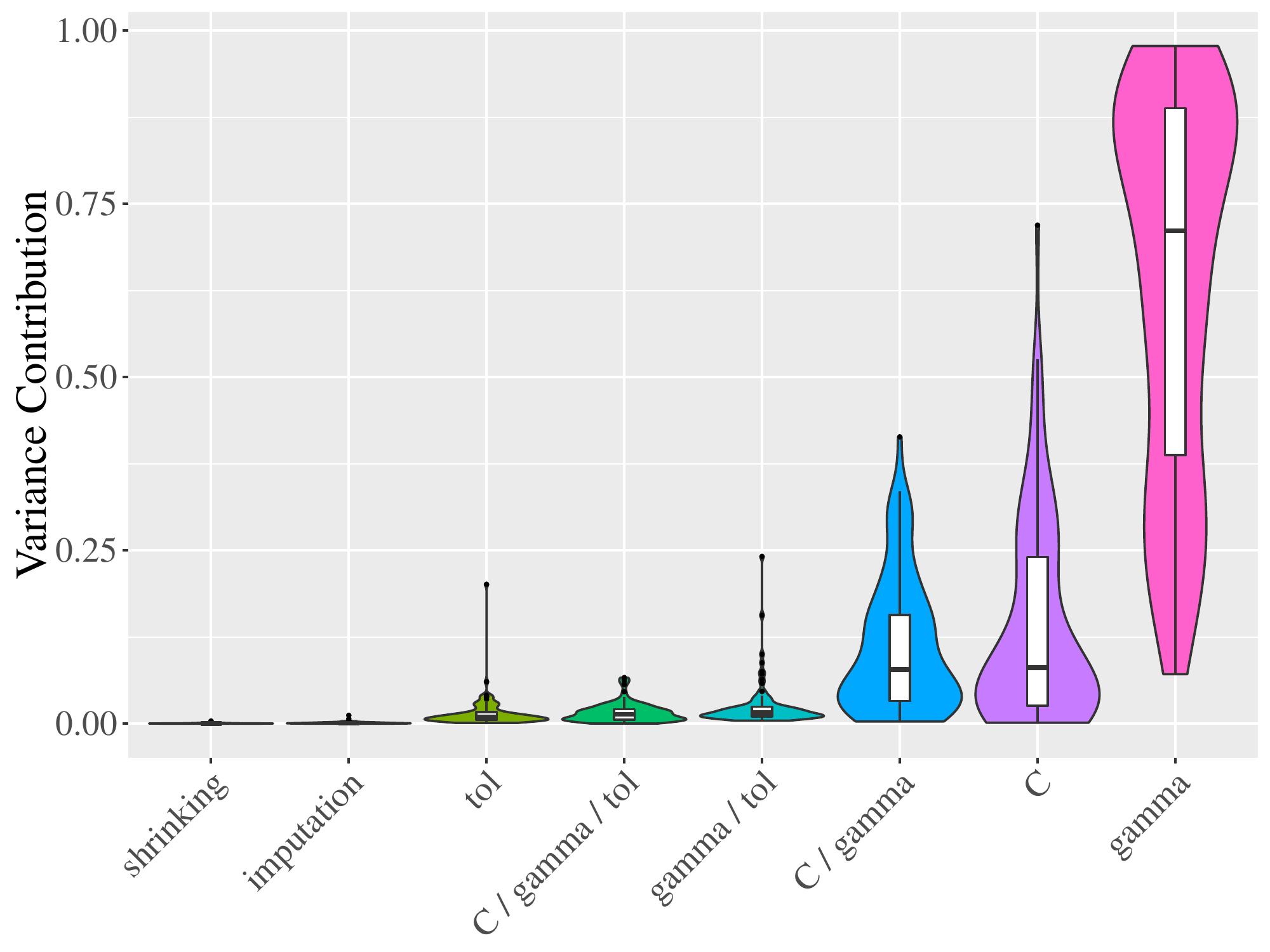}
      \label{fig:rbf_mc}
    }
    \subfigure[Random Search, excluding one parameter at a time] {
      \includegraphics[width=.75\columnwidth]{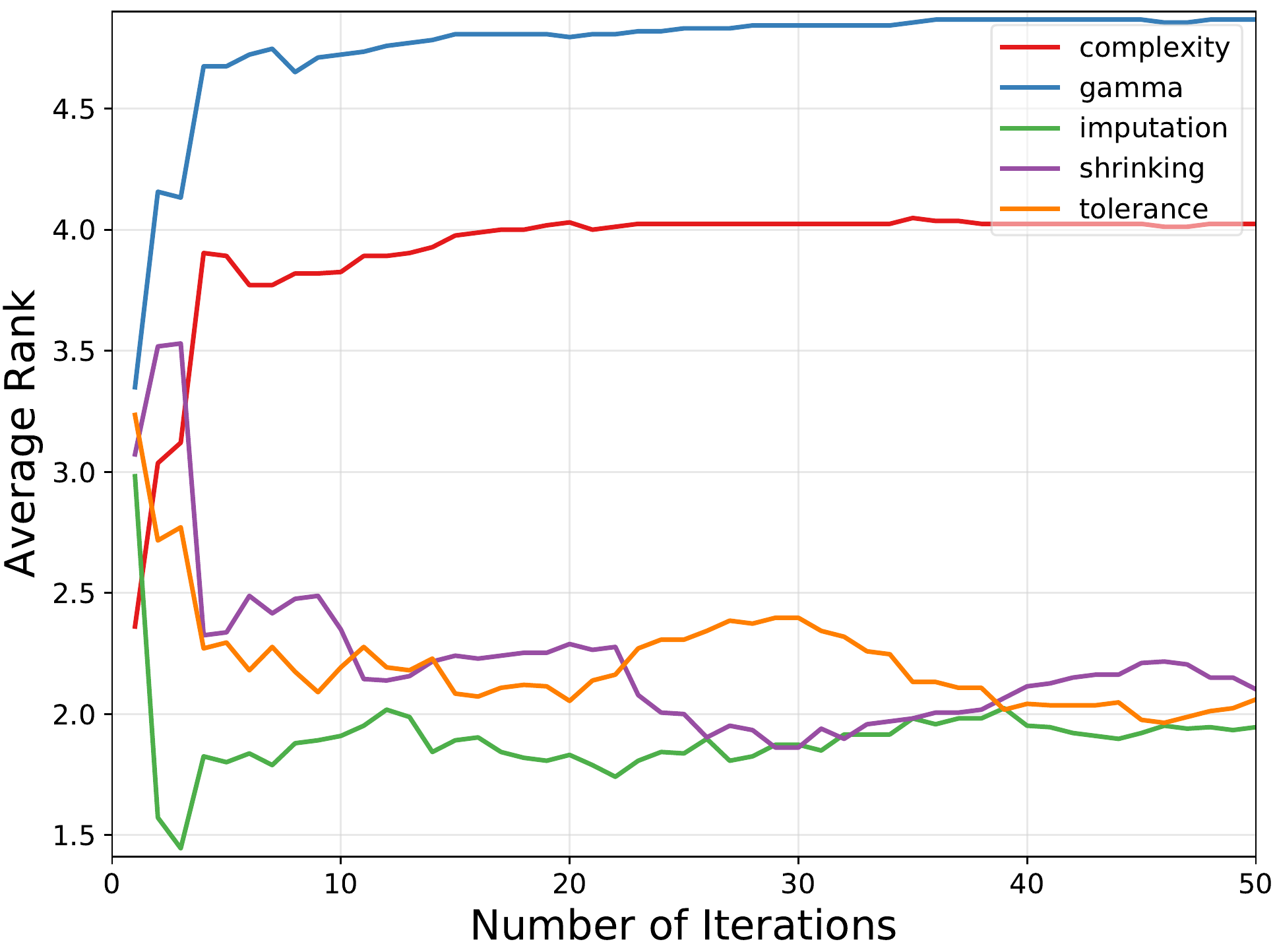}
      \label{fig:rbf_rs}
    }
    \subfigure[Ranked hyperparameter importance, $\alpha = 0.05$.]{
      \includegraphics[width=\columnwidth]{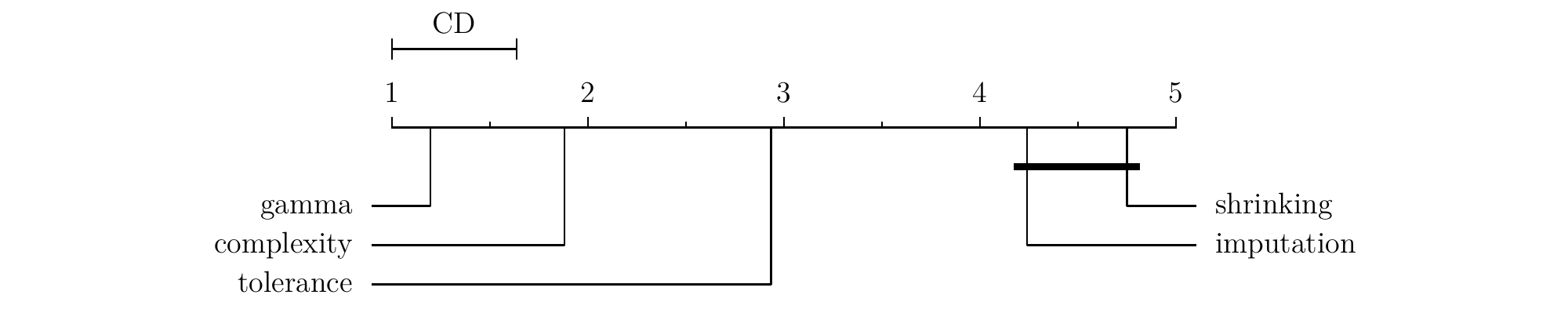}
      \label{fig:rbf_nemenyi}
    }
\caption{SVM (RBF kernel).\label{fig:rbf}}
  \end{center}
\end{figure}

We show the results for each classifier as a set of three figures. 
The top figure (e.g., Figure~\ref{fig:rbf_mc}) shows violinplots of each hyperparameter's variance contribution, across all datasets. The $x$-axis shows the hyperparameter $j$ under investigation, and each data point represents $\mathds{V}_{j} / \mathds{V}$ for one dataset; a high value implies that this hyperparameter accounted for a large fraction of variance on this dataset, and therefore would account for high accuracy-loss if not set properly. We also show for each classifier the three most important interaction effects between groups of hyperparameters. 

The middle figure (e.g., Figure~\ref{fig:rbf_rs}) shows the results of the verification experiment.
It shows the average rank of each run of random search, labeled with the hyperparameter whose value was fixed to a default value. 
A high rank implies poor performance compared to the other configurations, meaning that tuning this hyperparameter would have been important. 

The bottom figure (e.g., Figure~\ref{fig:rbf_nemenyi}) shows the result of a Nemenyi test over the average ranks of the hyperparameters (for details, see~\cite{Demsar2006}).
A statistically significant difference was measured for every pair of classifiers that are not connected by the horizontal black line. 
The interaction effects are left out to meet the independent input assumptions of the Nemenyi test. 

\begin{figure}[tb]
  \begin{center}
    \subfigure[Marginal contribution per dataset]{
      \includegraphics[width=.75\columnwidth]{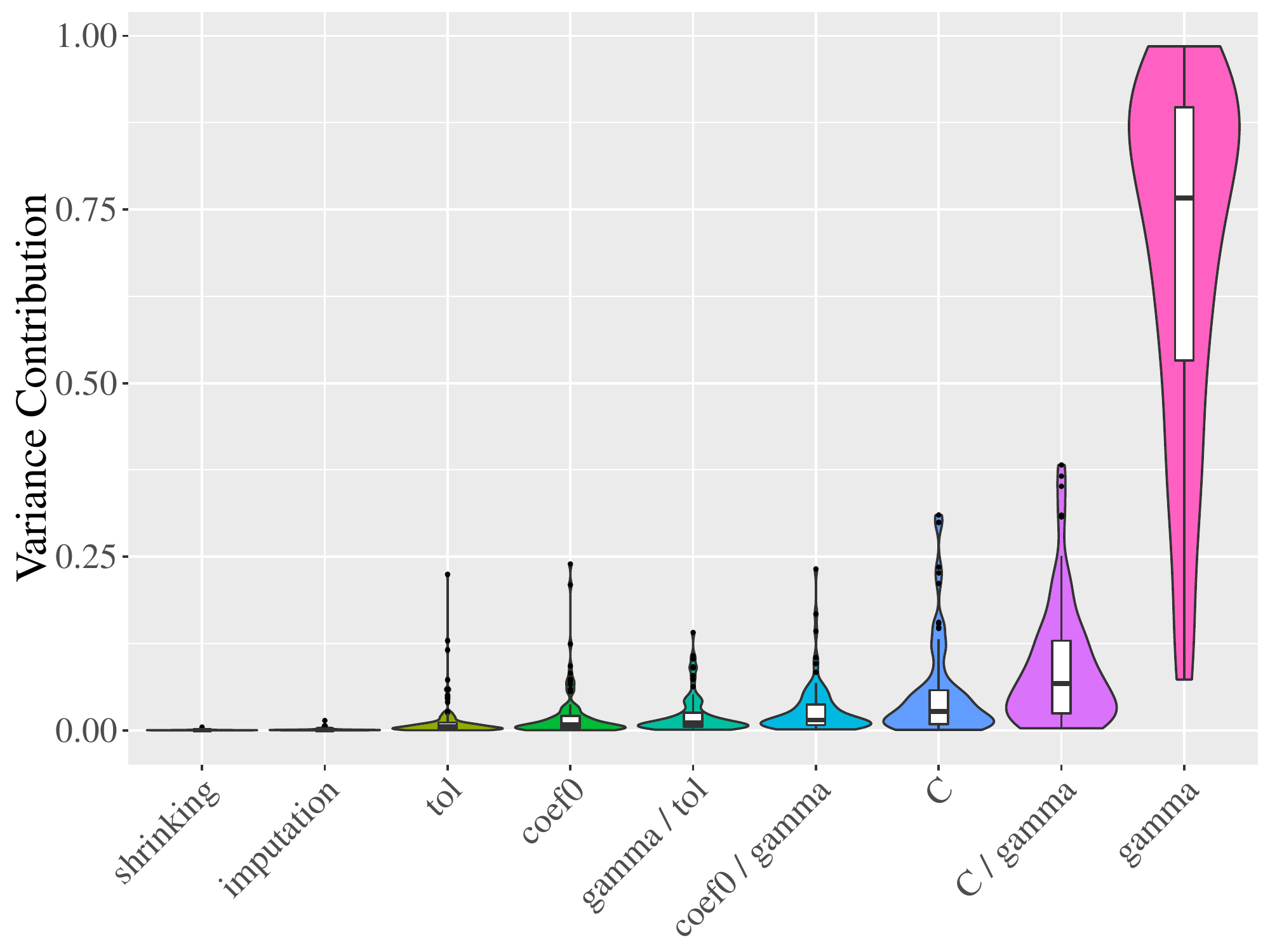}
      \label{fig:sigmoid_mc}
    }
    \subfigure[Random Search, excluding one parameter at a time] {
      \includegraphics[width=.75\columnwidth]{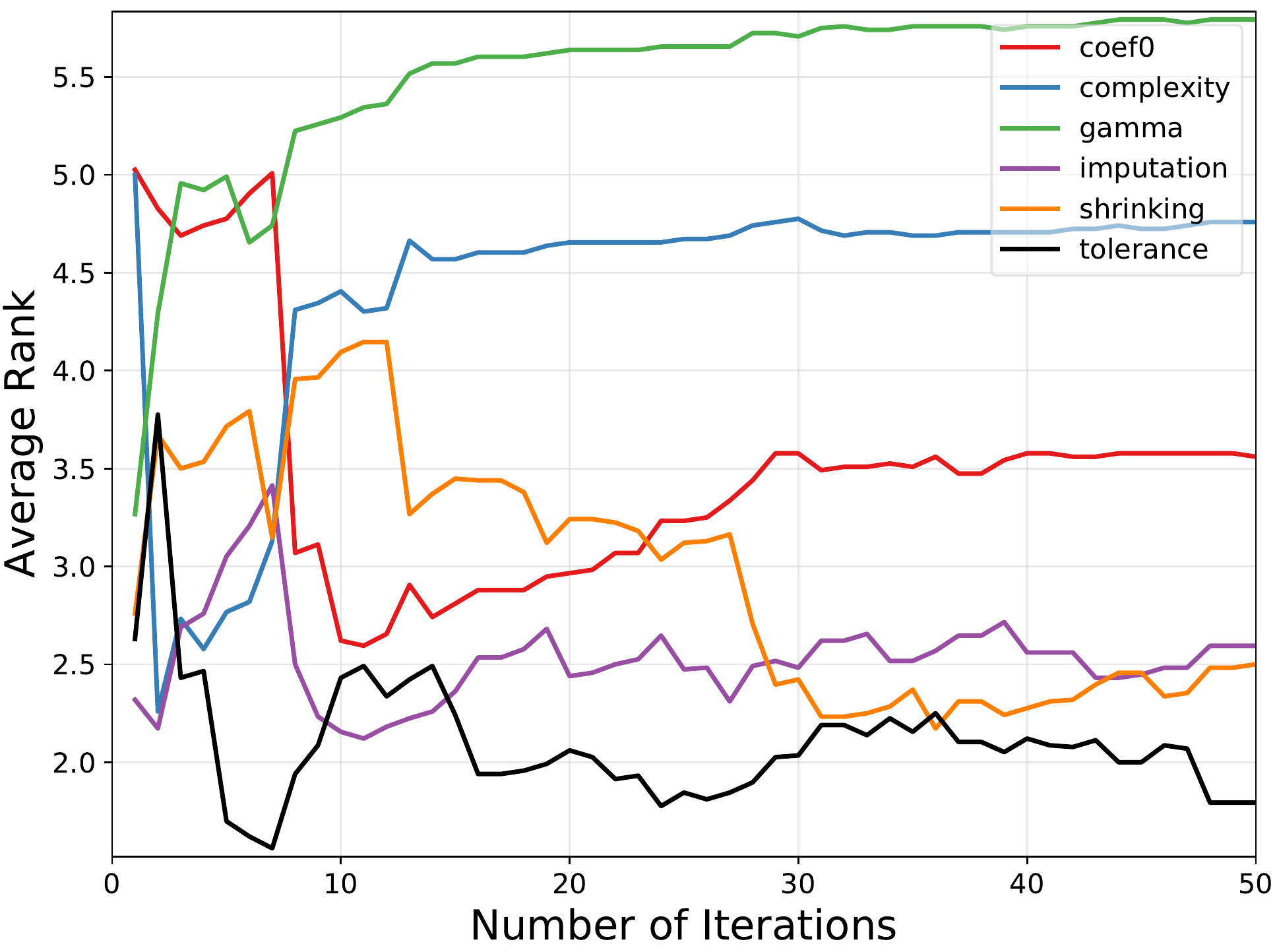}
    }
    \subfigure[Ranked hyperparameter importance, $\alpha = 0.05$.]{
      \includegraphics[width=\columnwidth]{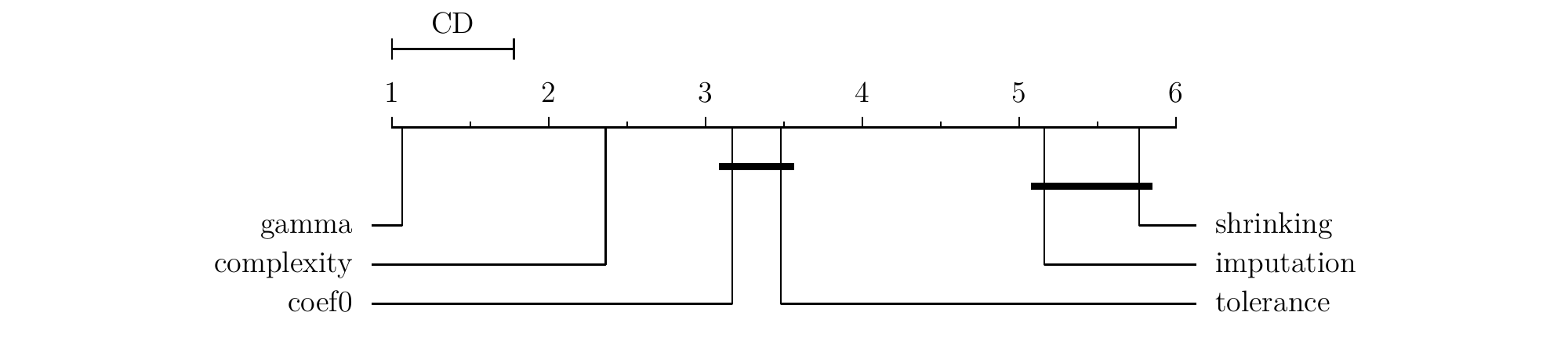}
    }
    \caption{SVM (sigmoid kernel).\label{fig:sigmoid}}
  \end{center}
\end{figure}

\paragraph{\bf SVM Results.}
We analyze SVMs with RBF and sigmoid kernels in Figures~\ref{fig:rbf} and \ref{fig:sigmoid}, respectively.


The results show a clear picture: The most important hyperparameter to tune in both cases was gamma, followed by complexity. Both of these hyperparameters were significantly more important than the others according to the Nemenyi test.
This conclusion is supported by the random search experiment: not optimizing the gamma parameter obtained the worst performance, making it the most important hyperparameter, followed by the complexity hyperparameter.
Interestingly, according to Figure~\ref{fig:sigmoid_mc}, when using the sigmoid kernel, the interaction effect between gamma and complexity was even more important than the complexity parameter by itself.

We note that while it is well-known that gamma and complexity are important SVM hyperparameters, to the best of our knowledge, this is the first study that provides systematic empirical evidence for their importance on a wide range of datasets. The fact that the proposed methods recovered these known most important hyperparameters also acts as additional verification that the proposed methodology works as expected.
The least important hyperparameter for the accuracy of SVMs was whether to use the shrinking heuristic. As this heuristic is intended to decrease computational resources rather than improve predictive performance, our data suggests that it is safe to enable this feature. 

\paragraph{\bf Random Forest Results.}

\begin{figure}[tb]
  \begin{center}
    \subfigure[Marginal contribution per dataset]{
      \includegraphics[width=.75\columnwidth]{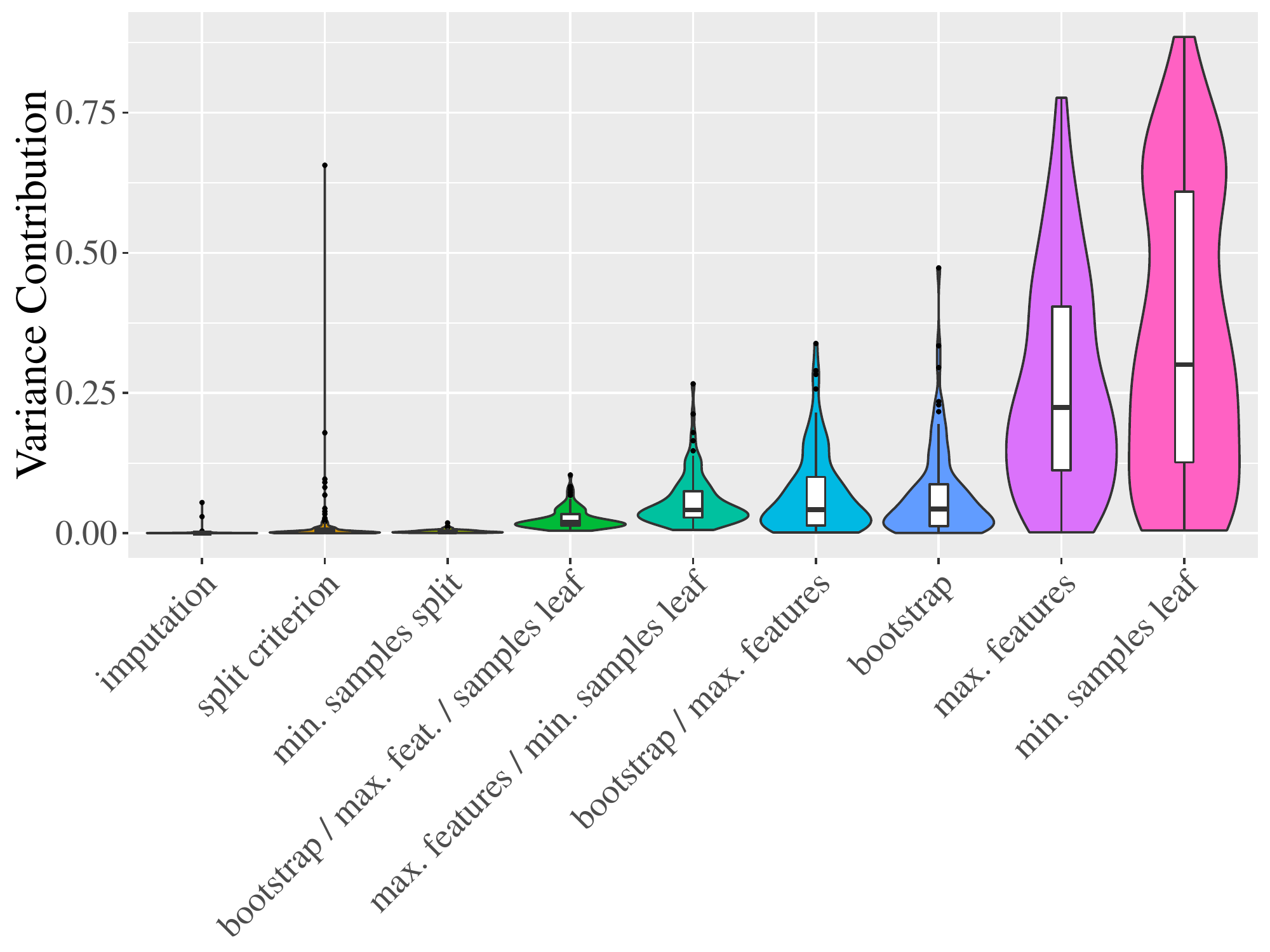}
      \label{fig:rf_mc}
    }
    \subfigure[Random Search, excluding one parameter at a time] {
      \includegraphics[width=.75\columnwidth]{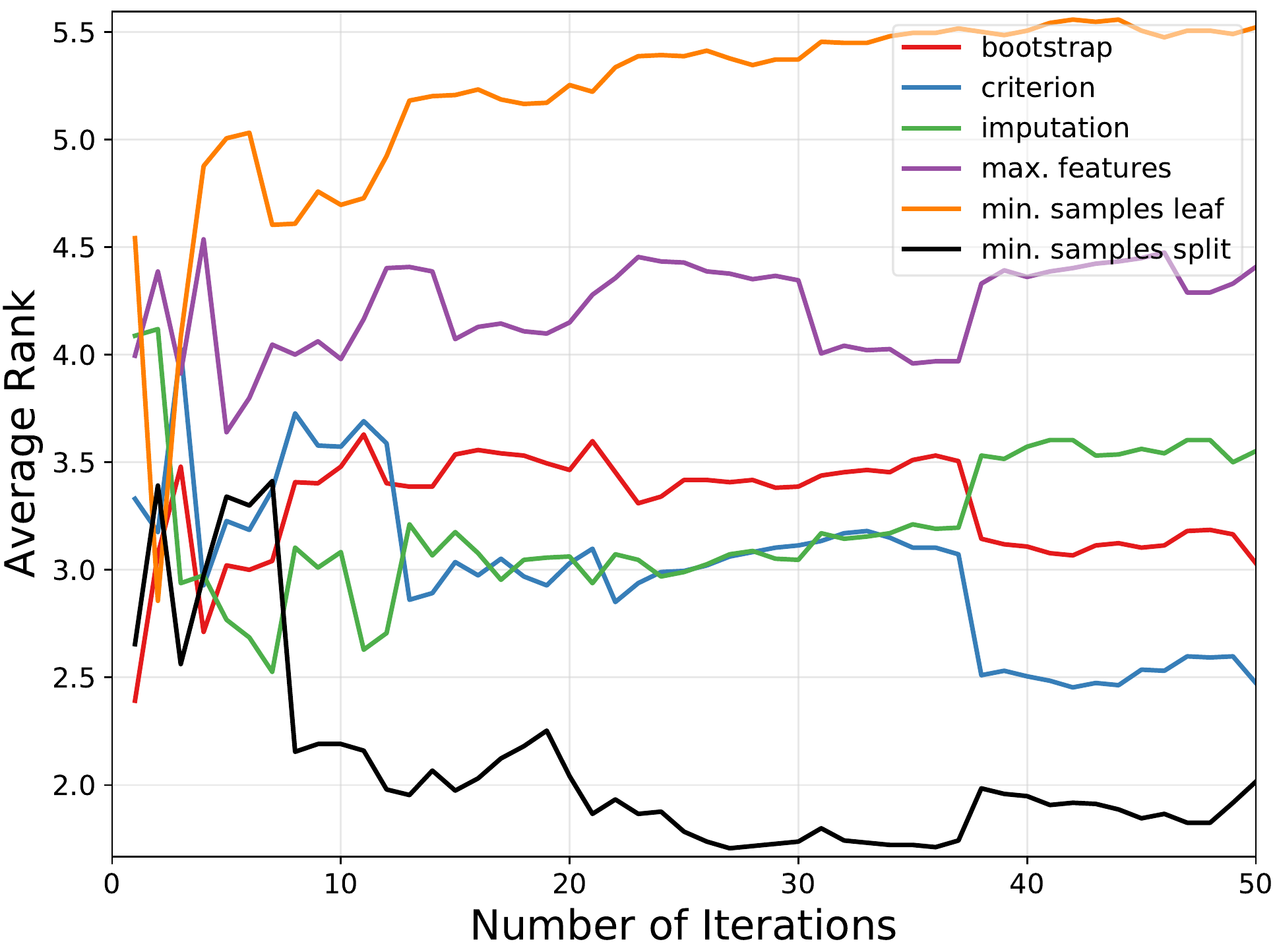}
    }
    \subfigure[Ranked hyperparameter importance, $\alpha = 0.05$.]{
      \includegraphics[width=\columnwidth]{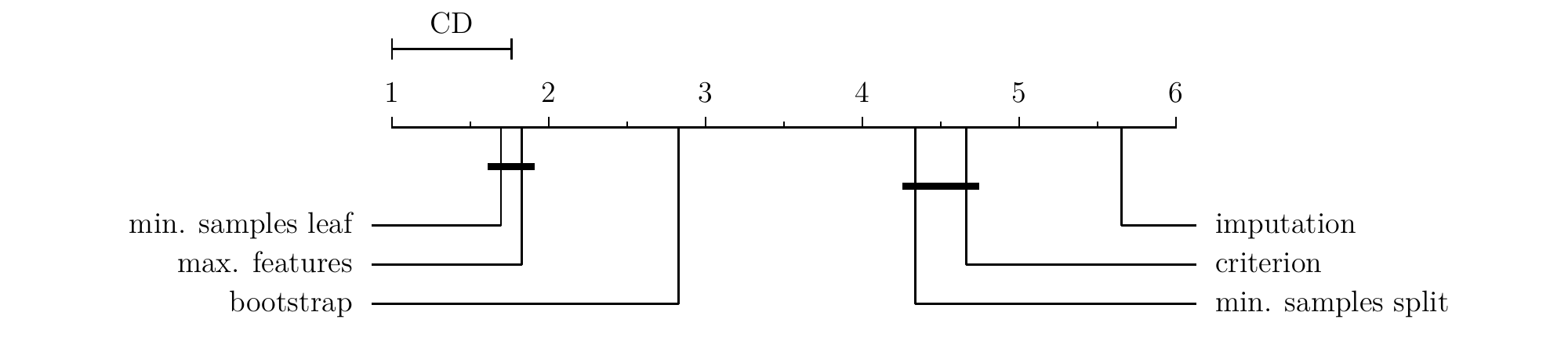}
    }
    \caption{Random Forest.\label{fig:random_forest}}
  \end{center}
\end{figure}

\begin{figure}[tb]
  \begin{center}
    \subfigure[Marginal contribution per dataset]{
      \includegraphics[width=.75\columnwidth]{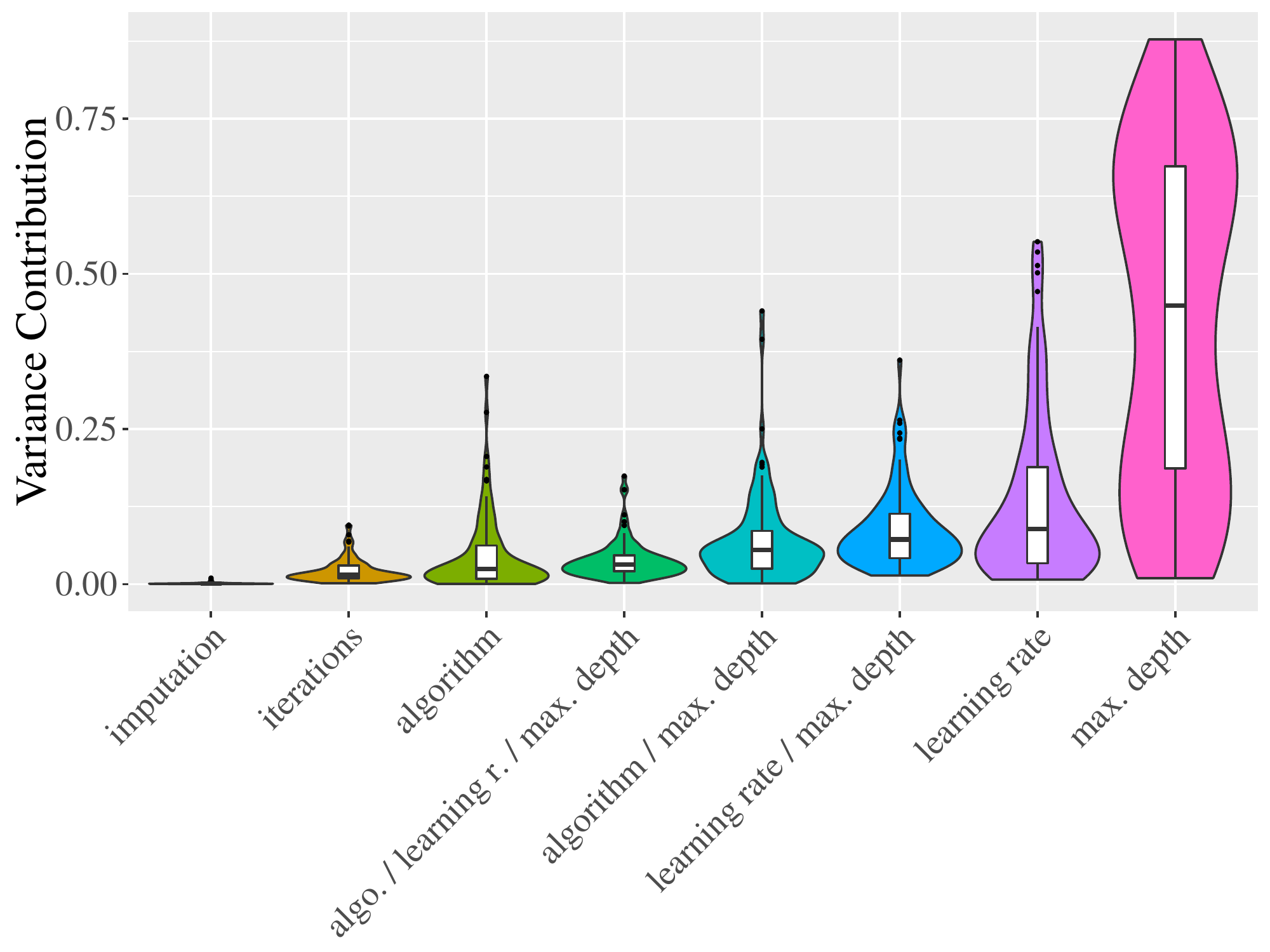}
      \label{fig:adaboostd_mc}
    }
    \subfigure[Random Search, excluding one parameter at a time] {
      \includegraphics[width=.75\columnwidth]{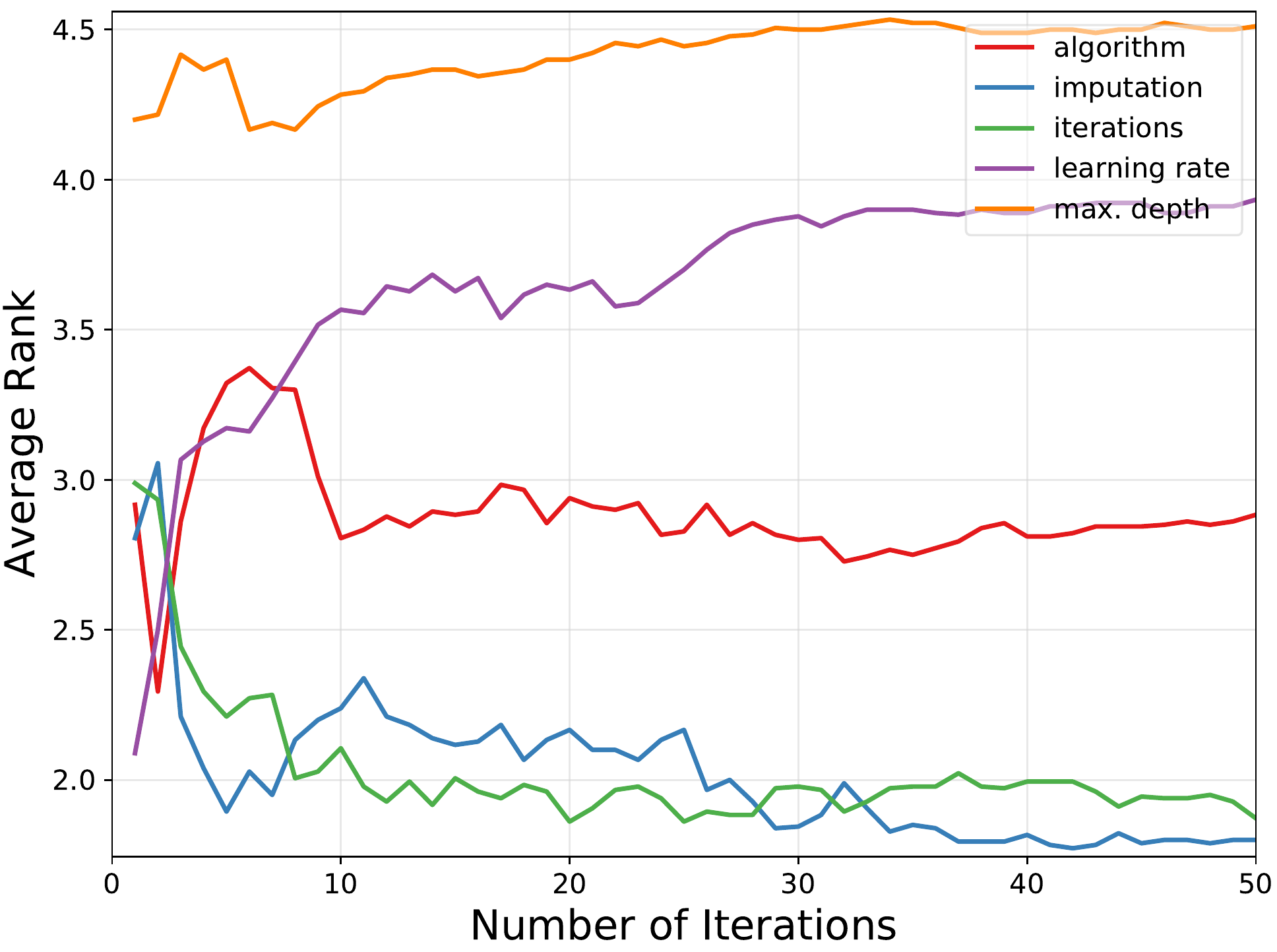}
    }
    \subfigure[Ranked hyperparameter importance, $\alpha = 0.05$.]{
      \includegraphics[width=\columnwidth]{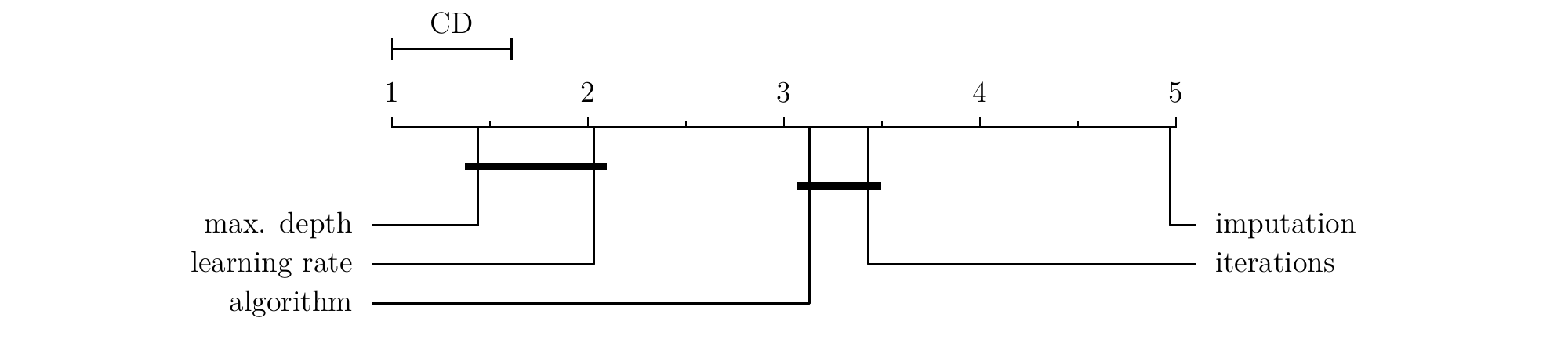}
    }
    \caption{Adaboost.\label{fig:adaboost}}
  \end{center}
\end{figure}

Figure~\ref{fig:random_forest} shows the results for random forests. 
The results reveal that most of the variance could be attributed to a small set of hyperparameters: the minimum samples per leaf and maximal number of features for determining the split were most important. Both of these hyperparameters were significantly more important than the others according to the Nemenyi test.
Only in a few cases, bootstrap was the most important hyperparameter (datasets `balance-scale', `credit-a', `kc1', `Australian', `profb' and `climate-model-simulation-crashes') and the split criterion only once (dataset `scene'). 
Again, the results from functional ANOVA agree with the results from the random search experiment and our intuition.
It is well-known that ensembles perform well when two conditions are met~\cite{Breiman2001,Hansen1990}:
\begin{enumerate*}[(i)]
    \item the individual models perform better than random guessing, and
    \item the errors of the individual models are uncorrelated. 
\end{enumerate*}
Both hyperparameters influence the variance among trees, uncorrelating their predictions. 

At first sight, the minimal samples per split and minimal samples per leaf hyperparameters seem quite similar, but at closer inspection they are not: logically, minimal samples per split is overshadowed by minimal samples per leaf. 

\paragraph{\bf Adaboost Results.} 
Figure~\ref{fig:adaboost} shows the results for Adaboost.  Again, most of the variance can be explained by a small set of hyperparameters, in this case the maximal depth of the decision tree and, to a lesser degree, the learning rate. Both of these hyperparameters were significantly more important than the others according to the Nemenyi test. There were only a few exceptions, in which the boosting algorithm was the most important hyperparameter (datasets `madelon', `diabetes' and `hill-valey'). The results were again confirmed by the verification experiment. 

One interesting observation is that, in contrast to other ensemble techniques, the number of iterations did not seem to influence performance too much. The minimum value ($50$) appears to already be large enough to ensure good performance, and increasing it does not lead to significantly better results. 

\paragraph{\bf General Conclusions.} 
For all classifiers, it appears that a small set of hyperparameters are responsible for most variation in performance. 
In many cases, this is the same set of hyperparameters across datasets. Knowing which hyperparameters are important is relevant in a variety of contexts, ranging from experimental setups to automated hyperparameter optimization procedures. 
Furthermore, knowing which hyperparameters are important is interesting as a scientific endeavor in itself, and can provide guidance for algorithm developers. 

Interestingly, the hyperparameter determining the imputation strategy did not seem to matter for any of the classifiers, even though the selected benchmarking suite contains datasets such as `KDDCup09 upselling', `sick' and `profb', all of which have many missing values. Imputation is clearly important (as classifiers do not function on undefined data), but which strategy to use for the imputation does not matter much according to the data.

We note that the results presented in this section do by no means imply that it suffices to tune just the set of most important hyperparameters. 
While the results by~\citeinline{Hutter2014} showed that this can indeed lead to faster improvements, they also indicated that it is still advisable to tune all hyperparameters when enough budget is available.
In the next experiment, as a complementary analysis, we will study which values are likely to yield good performance.

\section{Good Hyperparameter Values}
\label{sec:priors}
Now that we know which hyperparameters are important, the next natural question is which values they should be set to in order to likely obtain good performance. We now discuss the results of the experiment for answering this question.

\begin{figure*}[tb]
  \begin{center}
    \subfigure[RF: min. samples per leaf]{
      \includegraphics[width=.235\textwidth]{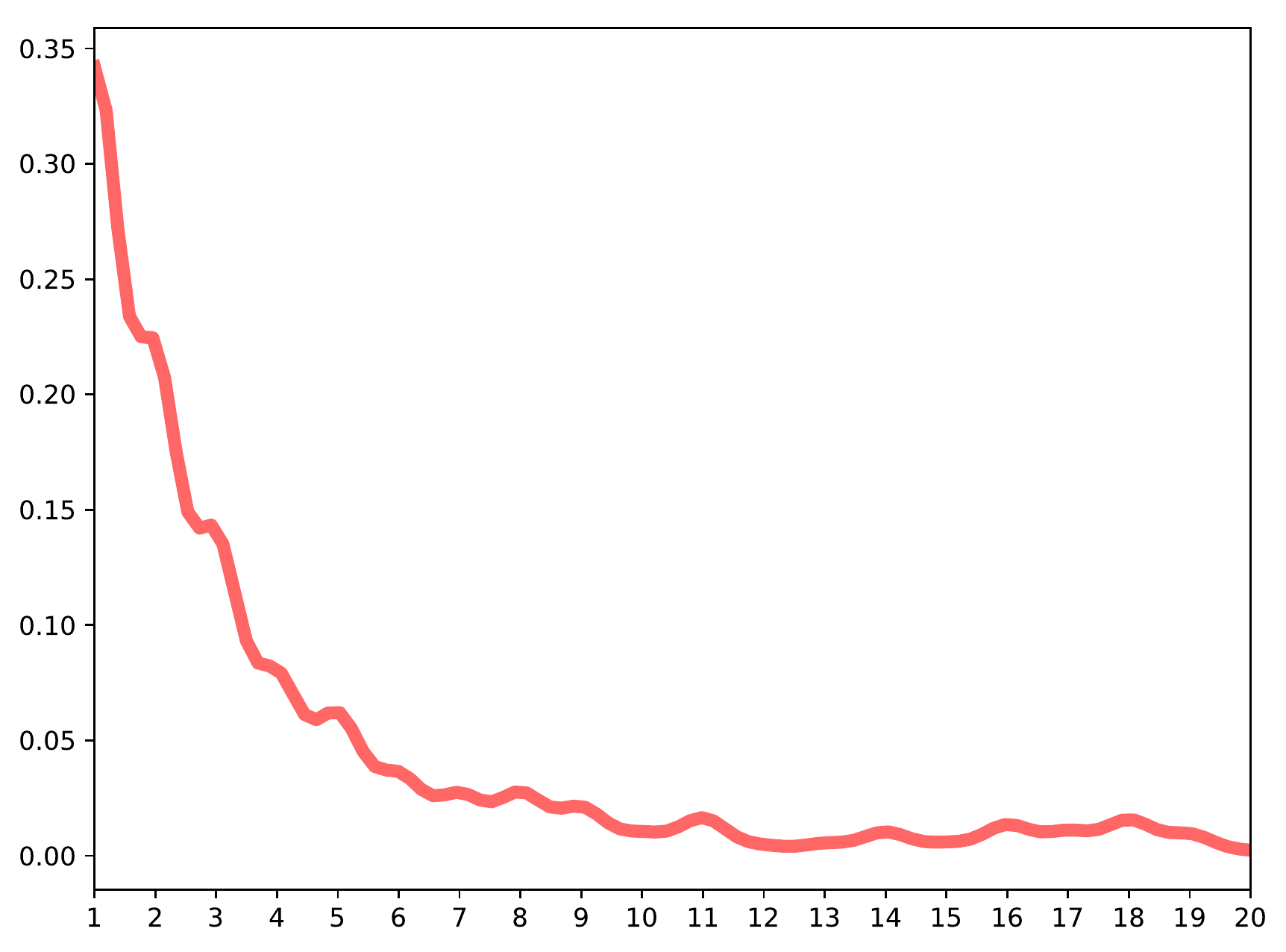}
    }
    \subfigure[Adaboost: max. depth of tree] {
      \includegraphics[width=.235\textwidth]{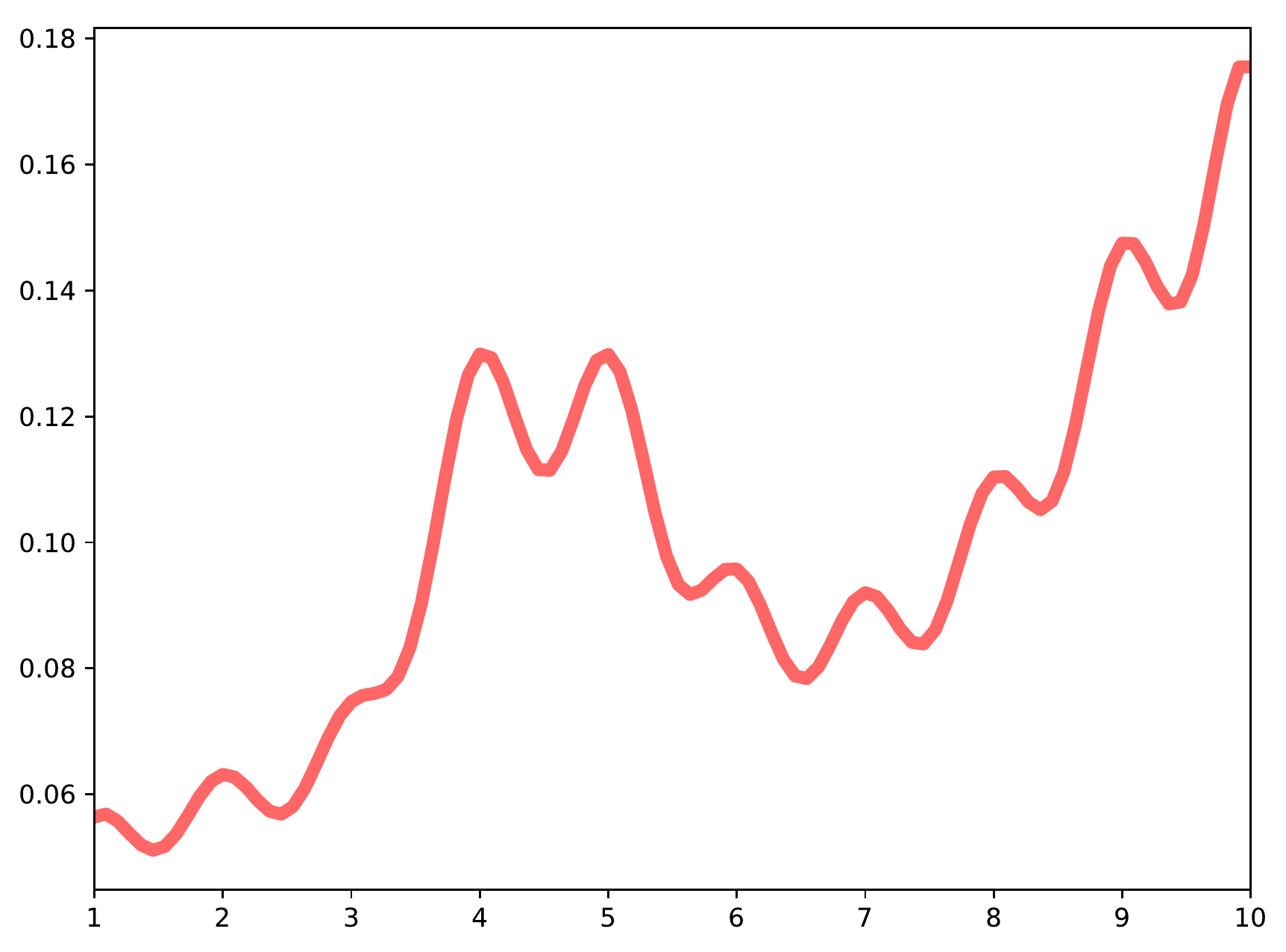}
    }
    \subfigure[SVM (RBF kernel): gamma] {
      \includegraphics[width=.235\textwidth]{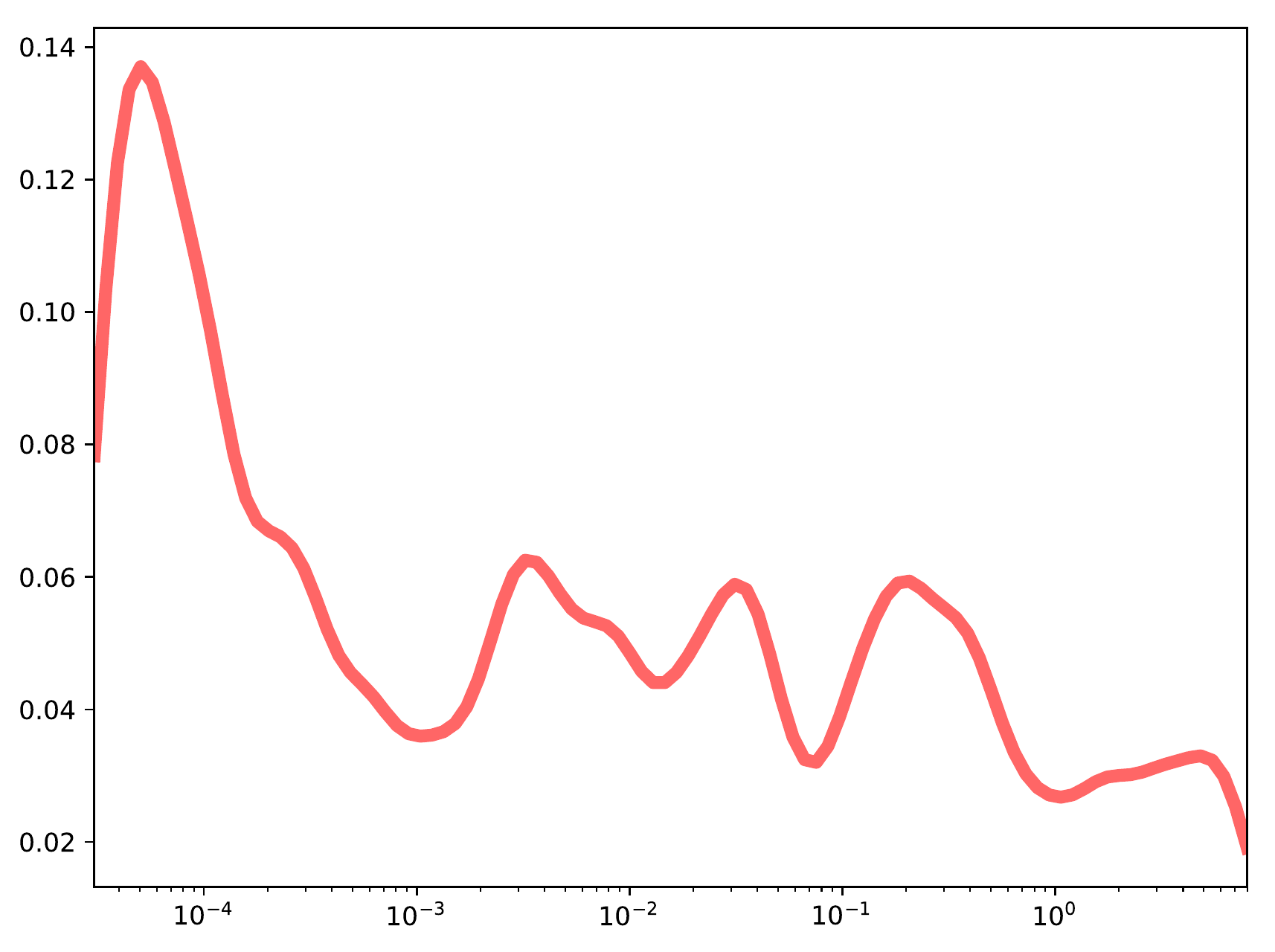}
    }
    \subfigure[SVM (sigmoid): gamma] {
      \includegraphics[width=.235\textwidth]{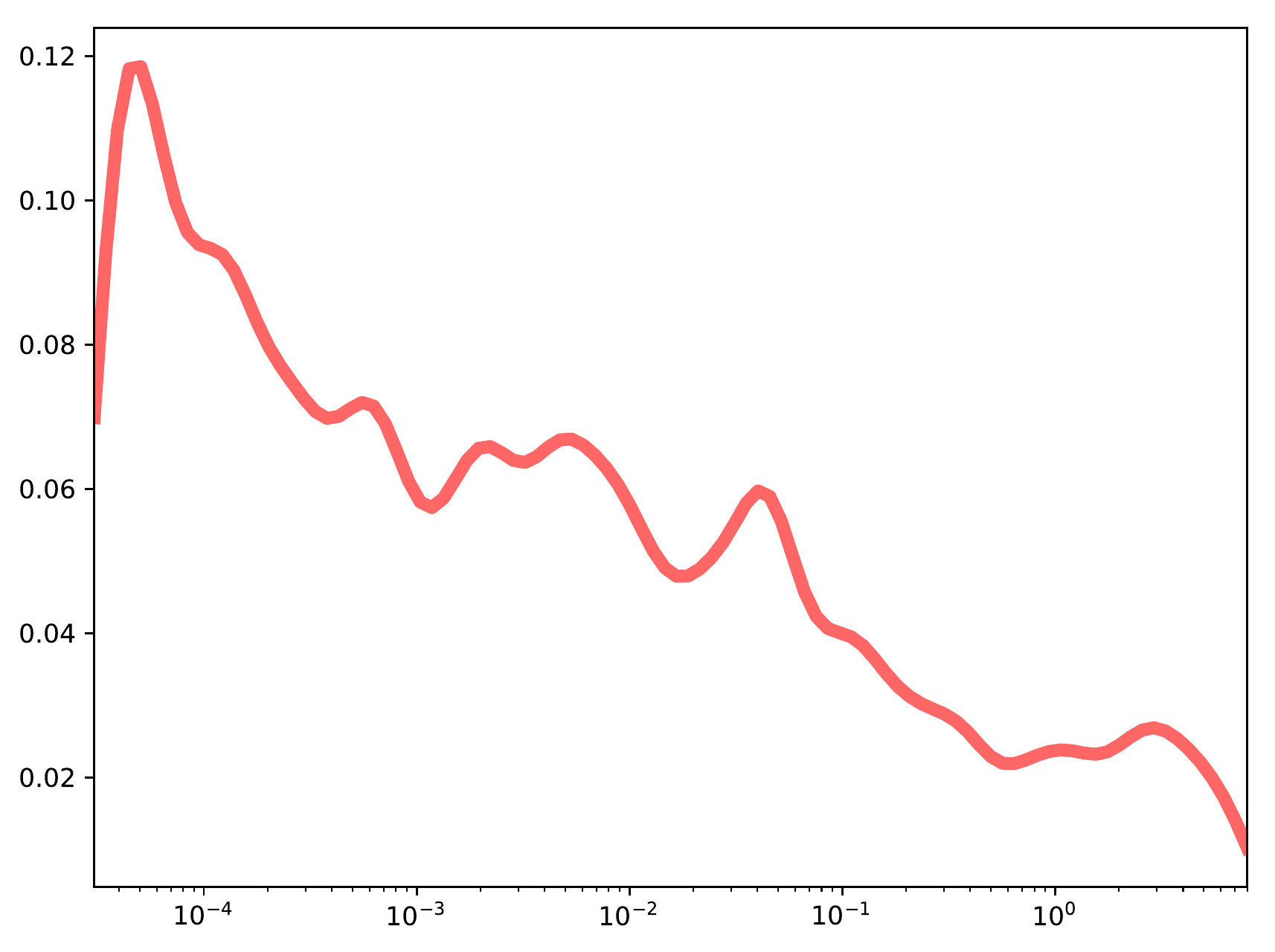}
    }
    \caption{Obtained priors for the hyperparameter found to be most important for each classifier. The $x$-axis represents the value, the $y$-axis represents the probability that this value will be sampled (integer parameters will be rounded). \label{fig:priors}}
  \end{center}
\end{figure*}

Figure~\ref{fig:priors} shows the kernel density estimators for the most important hyperparameters per classifier. 
It becomes clear that for random forests the minimal number of data points per leaf has a good default and should typically be set to quite small values. 
This is in line with the results reported by~\citeinline{Geurts2006} (albeit for the variant of `Extremely Randomized Trees'). 
Likewise, the maximum depth of the decision tree in Adaboost should typically be set to a large value. 
Both hyperparameters are commonly used for regularization, but the empirical data indicates that this should only be applied in moderation. 
For both types of SVMs, the best performance can typically be achieved with low values of the gamma hyperparameter.

\begin{figure}[tb]
  \begin{center}
    \subfigure[SVM (RBF)]{
      \includegraphics[width=0.22\columnwidth]{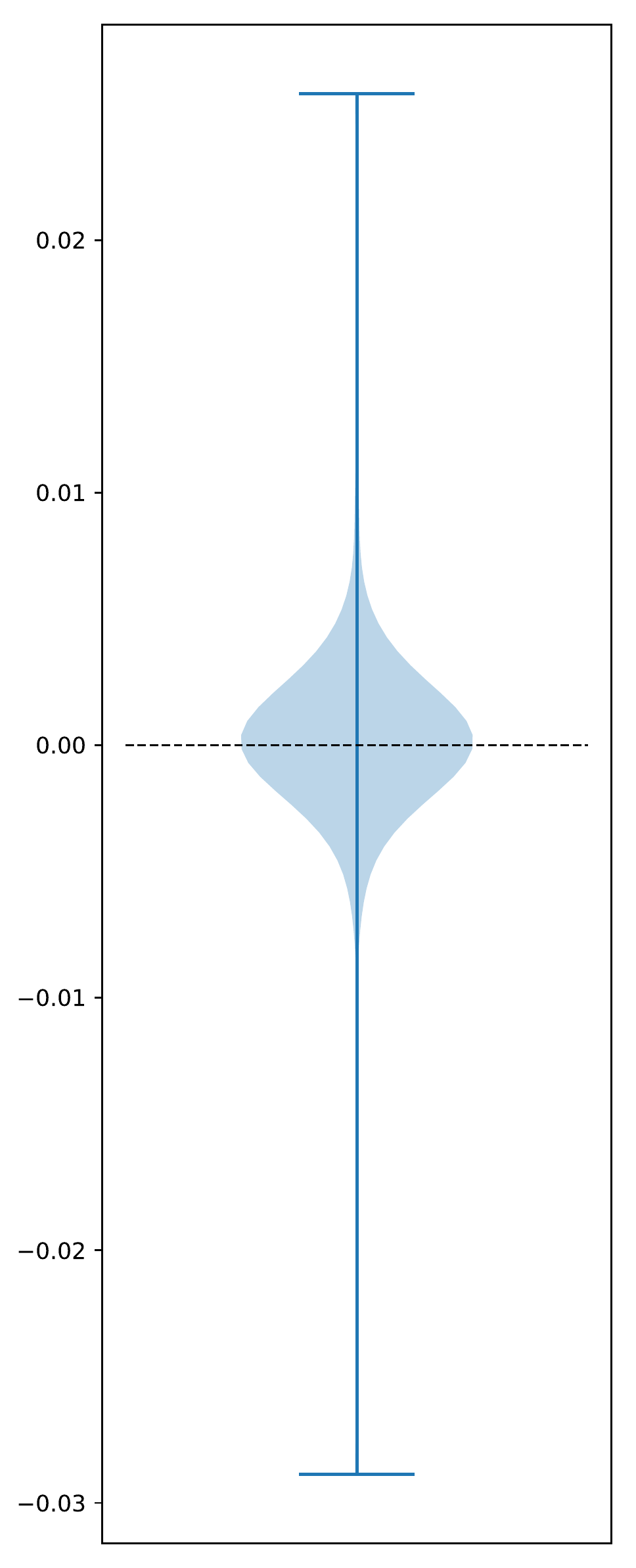}
    }
    \subfigure[Sigmoid]{
      \includegraphics[width=0.22\columnwidth]{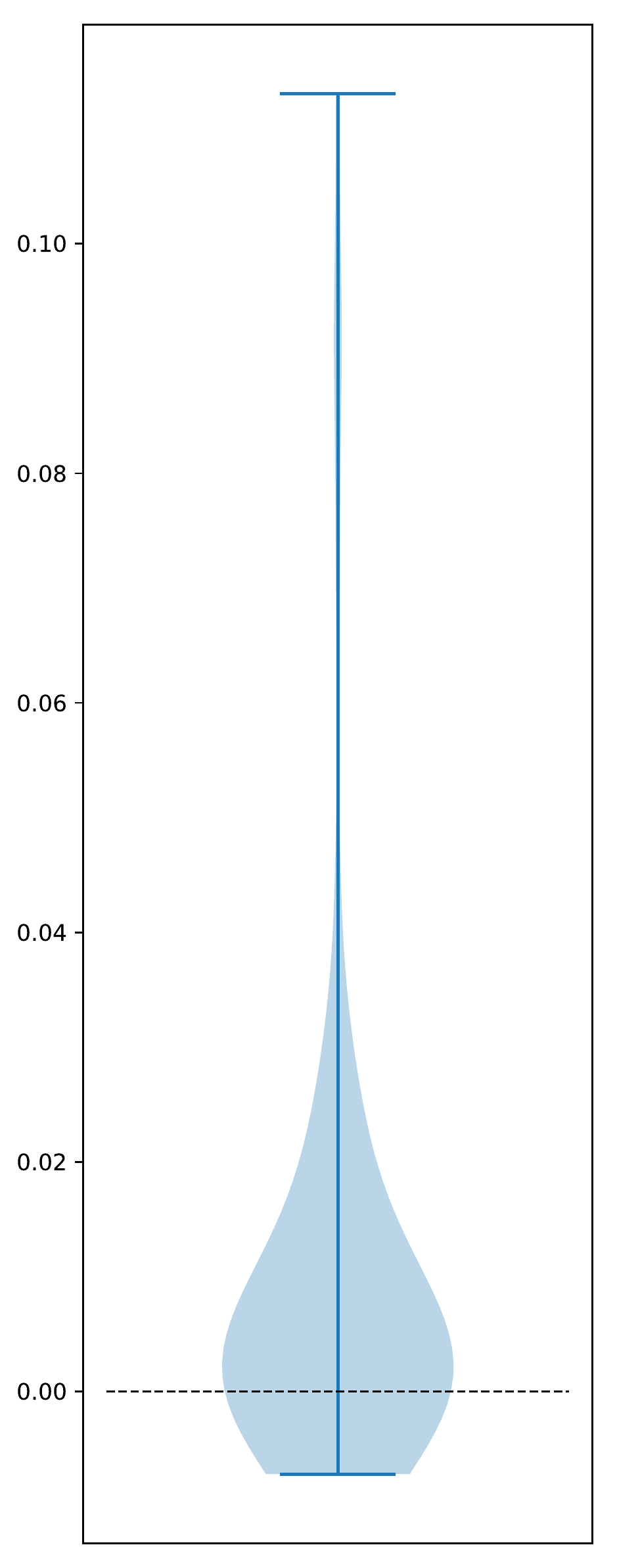}
    }
    \subfigure[Adaboost]{
      \includegraphics[width=0.22\columnwidth]{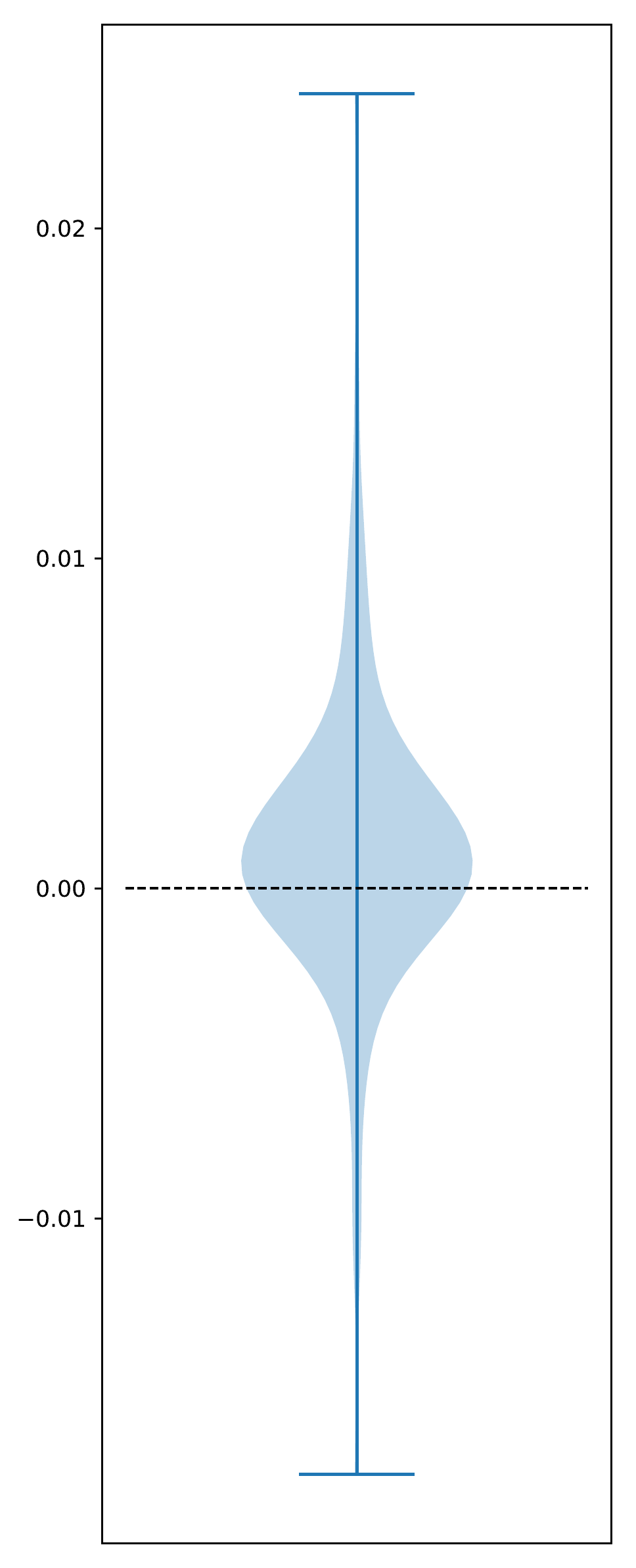}
    }
    \subfigure[RF]{
      \includegraphics[width=0.22\columnwidth]{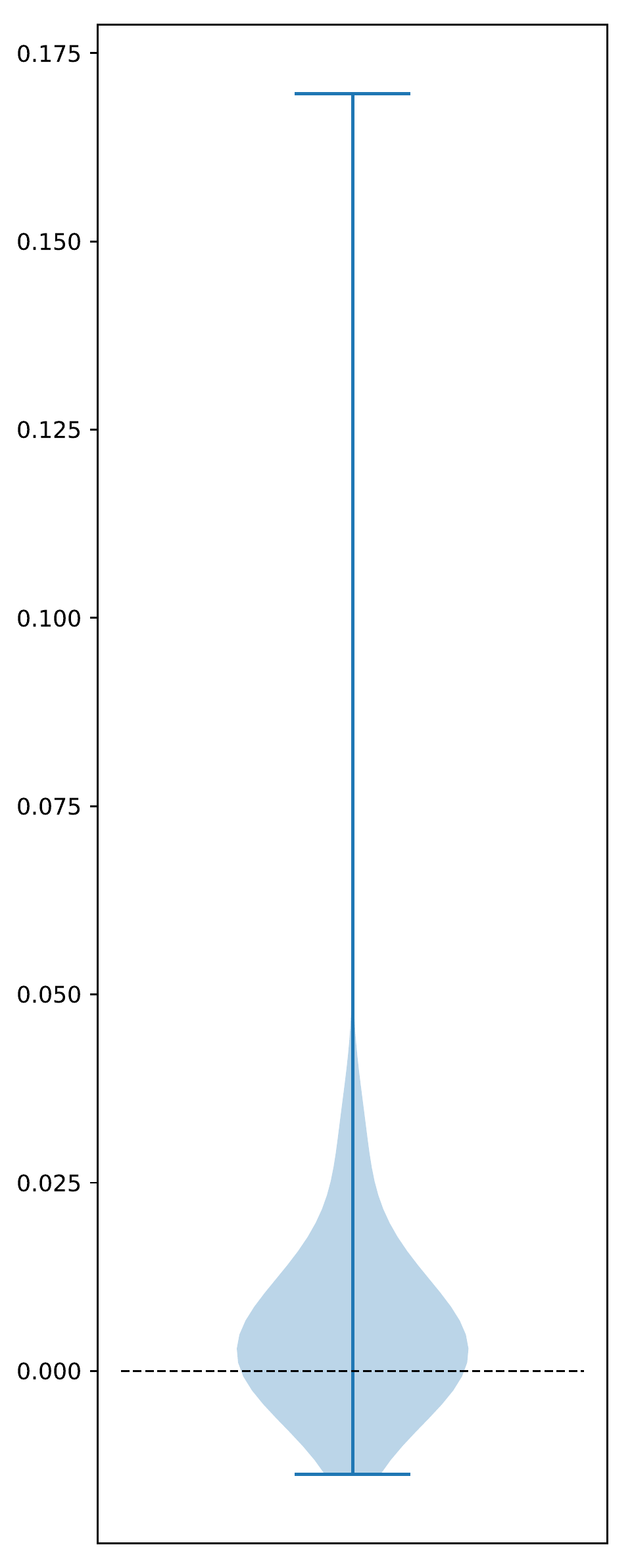}
    }
    \caption{Difference in performance between two instances of Hyperband, one sampling based on the obtained priors and one using uniform sampling. Values bigger than zero indicate superior performance for the procedure sampling based on the priors, and vice-versa. \label{fig:rs_priors}}
  \end{center}
\end{figure}

Next, we report the results of the experiment for verifying the usefulness of these priors in hyperparameter optimization. We do this in a leave-one-out setting: for each dataset under investigation, we build the priors based on the empirical performance data from the $99$ other datasets. 
Figure~\ref{fig:rs_priors} and Table~\ref{tab:nemenyi} report results comparing Hyperband with a uniform prior vs.\ the data-driven prior. Hyperband was ran with the following hyperparameters: 5 brackets, $s_{max} = 4$, $\eta = 2$ and $R = |\mathcal{D}^{(i)}|$ (the number of data points of dataset $\mathcal{D}^{(i)}$). Each optimizer was ran with $10$ different random seeds, and we report the average of their results.  

For each dataset, Figure~\ref{fig:rs_priors} shows the difference in predictive accuracy between the two procedures: values greater than $0$ indicate that sampling according to the data-driven priors was better by this amount, and vice versa. These per-dataset differences are aggregated using a violinplot. 
The results indicate that on many datasets the data-driven priors were indeed better, especially for random forests.

When evaluating experiments across a wide range of datasets, performance scales become a confounding factor. For example, for several datasets a performance improvement of $0.01$ already makes a great difference, whereas for others an improvement of $0.05$ is considered quite small. 
In order to alleviate this problem we conduct a statistical test, in this case the Nemenyi test, as recommended by~\citeinline{Demsar2006}. 
For each dataset, the Hyperband procedures are ranked by their final performance on the test set (the best procedure obtaining the lower rank, and an equal rank in case of a draw). 
If the ranks averaged over all datasets differ by more than a critical distance $CD$, the procedure with the lower rank performs statistically significant better. 

\begin{table}[tb]
  \begin{center}
    \caption{Results of Nemenyi test ($\alpha = 0.05, \mathit{CD} \approx 0.20$). We report ranks across $M$ datasets (max. 100), boldface the better approach (lower rank) and show whether the improvement is significant. \label{tab:nemenyi}}
    \begin{tabular}{l r r r r}
      \hline
      Classifier    &             $M$ & Uniform &      Priors & Sig. \\
      \hline
      random forest &           $100$ &  $1.72$ & $\bf{1.28}$ & yes \\
      Adaboost      &            $92$ &  $1.71$ & $\bf{1.29}$ & yes \\
      SVM (sigmoid) &            $86$ &  $1.73$ & $\bf{1.27}$ & yes \\
      SVM (RBF)     &            $89$ &  $1.60$ & $\bf{1.40}$ & yes \\
      \hline
    \end{tabular}
  \end{center}
\end{table}

The results of this test are presented in Table~\ref{tab:nemenyi}. 
We observe that the data-driven priors significantly improved performance over using uniform priors for all classifiers.\footnote{For the case of SVMs with RBF kernel, we note that the difference does not visually appear significant in Figure \ref{fig:rs_priors}, but using priors was better in 60\% of the datasets.}
The fact that the priors we obtained with a straightforward density estimator already yielded statistically significant improvements shows great promise. 
We see these simple estimators only as a first step and believe that better methods (e.g., based on traditional meta-learning and/or more sophisticated density estimators) are likely to yield even better results.

\section{Conclusions and Future Work}
\label{sec:conclusions}
In this work we addressed the questions which of a classifier's hyperparameters are most important, and what tend to be good values for these hyperparameters. 
In order to identify important hyperparameters, we applied functional ANOVA to a collection of $100$ datasets. 
The results indicate that the same hyperparameters are typically important for many datasets. 
For SVMs, the gamma and complexity hyperparameters are most important, for Adaboost the maximum depth and learning rate, and for random forests the minimum number of samples per leaf and maximum features available for a split. 
To the best of our knowledge, this is the first methodological attempt to demonstrate these findings across many datasets. 
In order to verify these findings, we conducted a large-scale optimization experiment, for each classifier optimizing all but one hyperparameter.
The results of this experiment are in line with the functional ANOVA results and largely agree with popular belief (for example, confirming the common belief that the gamma and complexity hyperparameters are the most important hyperparameters for SVMs). 
One surprising outcome of this analysis is that the strategy of data imputation hardly influences performance; investigating this matter further could warrant a whole study on its own, ideally leading to additional data imputation techniques. 

In order to determine which hyperparameter values tend to yield good performance, we fitted kernel density estimators to hyperparameter values that performed well on other datasets. This simple method already shows great promise based on the power of using data from many datasets: sampling from data-driven priors in hyperparameter optimization performed significantly better than sampling from a uniform prior.
We strove to keep all aspects of this work reproducible by anyone; we uploaded all the algorithm performance data to OpenML, including a Notebook for reproducing all results and figures in this paper.

In future work we plan to apply this analysis techniques to a wider range of classifiers.
While in this work we focused on more established types of classifiers to develop the methodology, quantifying important hyperparameters and good hyperparameter ranges of modern techniques, such as deep neural networks and extreme gradient boosting classifiers, could provide a useful empirical foundation to the field.
Furthermore, the developed methodology is by no means restricted to the classification setting; 
in future work, we plan to also apply it to regression and clustering algorithms. 
Finally, we aim to employ recent advances in meta-learning to identify similar datasets and base the priors only on these in order to yield dataset-specific priors for hyperparameter optimization.

~\\\noindent{}\textbf{Acknowledgements.}
We would like to thank Ilya Loshchilov for the valuable discussion that led to the methods we used for
\begin{enumerate*}[(i)]
    \item the verification of hyperparameter importance and 
    \item determining priors over good hyperparameter values.
\end{enumerate*}
This work has been supported by the European Research Council (ERC) under the European Union's Horizon 2020 research and innovation programme under grant no.\ 716721, the German state of Baden-W\"{u}rttemberg through bwHPC and the German Research Foundation (DFG) through grant no.\ INST 39/963-1 FUGG.

\bibliographystyle{ACM-Reference-Format}
\bibliography{hyperparameterimportance}

\end{document}